\documentclass{article}
\usepackage[]{iclr2025_conference}
\usepackage{amsmath,amsfonts,bm}

\def\eqref#1{equation~\ref{#1}}
\def\1{\bm{1}}

\DeclareMathAlphabet{\mathsfit}{\encodingdefault}{\sfdefault}{m}{sl}
\SetMathAlphabet{\mathsfit}{bold}{\encodingdefault}{\sfdefault}{bx}{n}

\newcommand{\E}{\mathbb{E}}

\DeclareMathOperator*{\argmax}{arg\,max}

\newcommand{\stack}[1]{\!\!\begin{array}{c}\scriptstyle #1\end{array}\!\!}

\newcommand{\smallfrac}[2]{{\textstyle\frac{#1}{#2}}}
\newcommand{\smallsum}[2]{ {\textstyle \sum\limits_{\scriptscriptstyle #1}^{\scriptscriptstyle #2}} }

\usepackage{hyperref}
\usepackage{url}

\usepackage[T1]{fontenc}    % use 8-bit T1 fonts
\usepackage{hyperref}       % hyperlinks
\usepackage{url}            % simple URL typesetting
\usepackage{booktabs}       % professional-quality tables
\usepackage{nicefrac}       % compact symbols for 1/2, etc.
\usepackage{microtype}      % microtypography
\usepackage{xcolor}         % colors

\usepackage{amsmath,amsfonts,bm}
\usepackage{graphicx}
\usepackage{multirow}
\usepackage{amssymb}
\usepackage{enumerate}
\usepackage{makecell}
\usepackage{algorithm, algorithmicx}
\usepackage[noend]{algpseudocode}
\usepackage{titlesec}
\usepackage{verbatim}

\usepackage[hashEnumerators,smartEllipses]{markdown}

\usepackage[sectionbib,globalcitecopy]{bibunits}

\def\V{\mathbb V\!}

\DeclareSymbolFont{matha}{OML}{txmi}{m}{it}% txfonts
\DeclareMathSymbol{\varv}{\mathord}{matha}{118}
\definecolor{nice_green}{rgb}{0, 0.5, 0}

\definecolor{wb}{rgb}{0, 0.5, 0}
\definecolor{removed}{rgb}{0.9, 0.9, 0.9}

\newtheorem{theorem}{Theorem}

\title{Epistemic Monte Carlo Tree Search}

\author{%
  Yaniv Oren, Viliam Vadocz, Matthijs T. J. Spaan \& Wendelin B{\"o}hmer \\
Delft University of Technology \\
2628 CD Delft, The Netherlands \\
\texttt{\{y.oren,v.vadocz,m.t.j.spaan,j.w.bohmer\}@tudelft.nl}
}

\author{%
  Yaniv Oren \\
  Delft University of Technology \\
2628 CD Delft, The Netherlands \\
\texttt{y.oren@tudelft.nl} \\
  \And Viliam Vadocz \\
  Delft University of Technology \\
2628 CD Delft, The Netherlands \\
  \And Matthijs T. J. Spaan \\
  Delft University of Technology \\
2628 CD Delft, The Netherlands \\
\texttt{m.t.j.spaan@tudelft.nl} \\
\And\hspace{26mm}
\And Wendelin B{\"o}hmer \\
  Delft University of Technology \\
2628 CD Delft, The Netherlands \\
\texttt{j.w.bohmer@tudelft.nl} \\
}

\iclrfinalcopy % Uncomment for camera-ready version, but NOT for submission.
\begin{document}

% Define the gaps between equations and text
\setlength{\abovedisplayskip}{4pt}
\setlength{\belowdisplayskip}{4pt}

% Define spacing before and after sections
\titlespacing\section{0pt}{12pt plus 4pt minus 2pt}{0pt plus 2pt minus 2pt}
\titlespacing\subsection{0pt}{12pt plus 4pt minus 2pt}{0pt plus 2pt minus 2pt}
\titlespacing\subsubsection{0pt}{12pt plus 4pt minus 2pt}{0pt plus 2pt minus 2pt}

\maketitle

\begin{abstract}
The AlphaZero/MuZero (A/MZ) family of algorithms has achieved remarkable success across various challenging domains by integrating Monte Carlo Tree Search (MCTS) with learned models.
Learned models introduce epistemic uncertainty, which is caused by learning from limited data and is useful for exploration in sparse reward environments.
MCTS does not account for the propagation of this uncertainty however.
To address this, we introduce Epistemic MCTS (EMCTS): a theoretically motivated approach to account for the epistemic uncertainty in search and harness the search for deep exploration.
In the challenging sparse-reward task of writing code in the Assembly language {\sc subleq}, AZ paired with our method achieves significantly higher sample efficiency over baseline AZ.
Search with EMCTS 
solves variations of the commonly used hard-exploration benchmark Deep Sea - which baseline A/MZ are practically unable to solve - much faster than an otherwise equivalent method that does not use search for uncertainty estimation,
demonstrating significant benefits from search for epistemic uncertainty estimation.
\end{abstract}

\section{Introduction}
\label{Introduction}
Many recent successes of reinforcement learning (RL) have been achieved by the model-based algorithm family of AlphaZero/MuZero \citep[A/MZ, ][]{AlphaZero, MuZero}.
A/MZ have outperformed humans in games, in tasks that traditionally relied on intricate human engineering \citep{mandhane2022muzero} and even made real world impact with the design of novel, more efficient algorithms \citep{AlphaTensor, AlphaDev} for day-to-day problems, a task that is often formulated as a challenging sparse reward environment.
When rewards are sparse, it is difficult to learn good policies without employing some form of \textit{deep exploration} to search for the rewards.
Deep exploration refers to the ability of the agent to direct itself towards novel transitions in the environment irrespective of how far away they are from the current state and promises up to exponential improvement in sample efficiency in sparse reward environments \citep{osband2018randomized}.

At the core of A/MZ is the combination of Monte Carlo Tree Search \citep[MCTS, ][]{swiechowski2023monte} with learned models of value and/or environment dynamics.
Learned models introduce \textit{epistemic uncertainty}, which refers to the uncertainty in the predictions of the model sourced in limited coverage of the state-action space during training \citep{hullermeier2021aleatoric}.
Accounting for epistemic uncertainty allows the agent to discern between predictions that are based on evidence (i.e.~the learned predictor was trained on this input), or based on generalization (i.e.~the learned predictor \textit{was not} trained on this input) and is useful for many purposes in online and offline RL.
Common uses range from reducing overestimation errors through pessimism in the face of uncertainty \citep[see][]{kumar2020conservative}, to directing deep exploration through optimism in the face of uncertainty \citep[see][]{unifying_ofu}.
Harnessing search for exploration is a popular approach in similar model based algorithms, such as Dreamer \citep[][]{sekar2020planning}.

MCTS, however, was designed for search with the true dynamics model and without a value model, and as a result, does not account for epistemic uncertainty introduced from learning the models.
For this reason, A/MZ cannot harness MCTS for deep exploration, nor benefit from the epistemic uncertainty associated with the predictions of the search tree in other ways.
In this work we aim to address both, with three main contributions.
Practical and theoretically motivated methods to 
(i) harness MCTS and epistemic uncertainty for upper-confidence-bound-based deep exploration \citep[][]{jin2018q} and 
(ii) propagate the epistemic uncertainty from learned models of value and/or dynamics during search, which we call Epistemic MCTS (EMCTS). 
(iii) A parallelized implementation in JAX \citep{jax2018github} of EMCTS paired with an AZ agent and an environment implementing the Assembly language {\sc subleq} 
 \citep{subleq}\footnote{
 Our implementation, inspired by \href{https://jaredkrinke.itch.io/sic-1}{{\sc sic-1}} which is an open source game demonstrating {\sc subleq}, is available at \url{https://github.com/emcts/e-alphazero}.
 }.
We find that the propagation of epistemic uncertainty in EMCTS is very similar to that of value in MCTS.
As search with MCTS improves the value estimates at the root, we hypothesize that search with EMCTS similarly improves the epistemic uncertainty estimates at the root, resulting in more accurate UCBs and more sample efficient exploration compared to a method that does not rely on search but is otherwise equivalent.

We evaluate EMCTS in the challenging, similar to real-world applications and sparse-reward task of programming in {\sc subleq}, as well as in the commonly used hard-exploration benchmark Deep Sea \citep{osband2019behaviour}.
Our method is able to find correct programs for a harder programming task in a much smaller number of samples than the AZ baseline.
In the Deep Sea benchmark, our method demonstrates deep exploration by solving stochastic and deterministic reward variations of the task, both of which baseline A/MZ is unable to solve in a reasonable number of samples.
In addition, EMCTS significantly outperforms an ablation that does not rely on search for epistemic uncertainty estimation but is otherwise equivalent, demonstrating significant advantages from search for uncertainty estimation.

\section{Background}
\label{background}

In RL, an agent learns a behavior policy through interactions with an environment.
The environment is represented by a Markov Decision Process \citep[MDP,][]{bellman1957markovian} $ \mathcal{M} = \langle \mathcal{S}, \mathcal{A}, \rho, \mathcal{R}, P,
\gamma \rangle $, where $ \mathcal{S} $ is a set of states, $ \mathcal{A}$ a set of actions, $ \rho $ the initial state distribution,
$ \mathcal{R}: \mathcal{S} \times \mathcal{A} \to \mathbb{R} $ a bounded possibly stochastic reward function, and $ P $ is a transition distribution such that $P(s' | s, a)$ specifies the probability of transitioning from state $s$ to state~$s'$ after executing action $a$.
The objective $ J_\pi $ of the agent is to find a policy $\pi(a|s)$, specifying the probability of selecting action $a$ in state $s$, that maximizes the \textit{expected discounted return}, also called value $V^\pi$, from the starting state distribution $\rho$: 
\begin{align}
    J_\pi = \E[V^\pi(s_0) | \stack{s_0 \sim \rho}\,] = 
\E \Big[{\textstyle\sum\limits_{t=0}^{H-1}}
\gamma^{t}\mathcal{R}(s_t, a_t) \Big| \stack{s_0 \sim \rho, s_{t+1} \sim P(s_t, a_t), a_t \sim \pi(s_t)}\!
\Big].
\end{align}
The discount factor $0 < \gamma < 1$ is used in infinite-horizon MDPs,
i.e.~$H=\infty$,
to guarantee that the values remain bounded, and is commonly used in RL for training stability.
A state-action \textit{Q-value function} is also often used:
$Q^\pi(s, a) = \E[\mathcal{R}(s, a) + \gamma V^\pi(s') | \stack{s' \sim P(s, a)}\!]$.
We denote the value of the optimal policy $\pi^*$ with $ V^*(s) = \max_\pi V^\pi(s), \forall s \in \mathcal{S} $.
In offline RL, the agent must maximize this objective given a static dataset.
In model-based RL (MBRL) the agent uses a model of the dynamics of the environment $(P, \mathcal{R})$ to optimize its policy, often through planning \citep[]{moerland2023model}. 
The dynamics are either learned from interactions \citep[e.g.,~in MZ,][]{MuZero} or provided \citep[e.g.,~in AZ,][]{AlphaZero}.
In Deep MBRL the agent utilizes deep neural networks \citep[DNN,][]{goodfellow2016deep} to approximate any of the value, policy, reward and transition functions.

\subsection{Monte Carlo Tree Search}
\label{MCTS}
The MCTS algorithm constructs a tree with the current state $s_t$ at its root to estimate the objective: $\argmax_{a}  \max_\pi Q^\pi(s_t, a)$ \citep[]{MCTS_survey}, by iteratively performing selection, expansion, simulation and backup.
At each iteration $i$ of the algorithm a trajectory in the tree is selected using a tree search policy such as UCT~\citep[][]{UCT}:
\begin{align}
    a^i_k \;=\; 
    UCT_i(s_k) \;=\; 
    \argmax_{a \in A}
    q^i(s_k, a) + C_{UCT}  {\textstyle\sqrt{\frac{2 \log (\sum_{a'}\! N(s_k, a'))}{N(s_k, a)}}}
    \label{UCT},
\end{align}
where $N(s_k, a)$ denotes the number of times action $a$ has been selected in node $s_k$, $C_{UCT} > 0$ trades off exploration of new nodes in the tree with maximizing observed return and $ q^i(s_k, a) $ is the averaged return observed for this state action up to step $i$.
Modern algorithms (such as A/MZ) use instead variations of PUCT \citep{rosin2011multi}:
\begin{align}
    a^i_k \;=\; 
    PUCT_i(s_k) \;=\; 
    \argmax_{a \in A}
    q^i(s_k, a) + \pi(a | s_k) \,C_{PUCT} {\textstyle\frac{ \sqrt{\sum_{a'}\! N(s_k, a')}}{1 + N(s_k, a)}}, 
    \label{eq:PUCT}
\end{align}
with some learned prior-policy $\pi$.
When selection step $i$ arrives at a leaf $s^i_T$ the node is expanded with a value estimate $v(s^i_T)$.
MCTS propagates the return $\nu^i$ of planning step $i$ back along the planning trajectory 
$\{s^i_{j}, a^i_{j}\}_{j=0}^{T}$:
\begin{align}
     \nu^i(s_k, a^i_k) = \sum_{j = 0}^{T-1}\gamma^j r(s^i_{k+j}, a^i_{k+j}) + \gamma^{T}v(s^i_{T}), 
     \quad q^i(s_k,a) = \frac{1}{N(s_k,a)} \sum_{j=1}^{i} \nu^j(s_k,a)
     \label{eq:backup}
\end{align}
where $\textstyle{a^i_{k+j} = P/UCT_i(s^i_{k+j})} $ and $  \textstyle{s^i_{k+j + 1} = f(s_{k+j}^i, a^i_{k+j})}$ 
for a deterministic transition function\footnote{
    For notational simplicity we assume here a deterministic transition function $f$, so that nodes $s^i_{k+j}$ correspond to individual states. It is also possible to use a stochastic transition model $P$, where nodes correspond to distributions of states, which are sampled by the selection step. The above equations remain the same, but the state corresponding to $s^i_{k+j}$ is effectively a random variable.
} $f: \mathcal{S} \times \mathcal{A} \to \mathcal{S} $ and a mean-reward function $ r(s,a) = \E[\mathcal{R}(s,a)]$.
In the original MCTS, the dynamics model $ m = (f,r) $ is provided to the agent and assumed to be correct for the true MDP $\mathcal{M}$ and the value estimate $ v(s^i_{T}) $ is computed using Monte-Carlo (MC) rollouts with $m$.
In AZ the value function \citep{AlphaZero} and in MZ all functions $f,r,v$, are learned from interactions with the environment 
and thus their predictions are uncertain outside of the training set.
The prior-policy $ \pi $ is trained with cross-entropy loss on targets extracted from the root of the tree.
With Reanalyze \citep{schrittwieser2021online}, which uses search to generate fresh value and policy targets from off-policy data, A/MZ are able to learn off-policy.

\subsection{Quantifying Uncertainty in Deep Reinforcement Learning}
\label{background_uncertainty}
Quantifying uncertainty in deep learning is an active field of research \citep[see][]{hullermeier2021aleatoric, lockwood2022review}.
In this work, we take the common approach for quantifying epistemic uncertainty as \textit{the variance in a random variable that approximates predictions that are consistent with a dataset of observed interactions} $\mathcal{D}$. We model a learned function $ \hat r$, trained to approximate a true function $r$ on data $\mathcal{D}$, $ \hat r(s,a) \approx r(s,a) $, as a random variable with respect to the data: $ \hat r(s,a) | \mathcal{D} := \hat R(s,a) $.
We assume unbiased approximation
$\E [\hat R(s,a)] := r(s,a)$, 
and use the variance $ \V\,[\hat R(s,a)]$ to quantify the epistemic uncertainty.
In discrete state-action spaces, the epistemic uncertainty in a reward or a transition can be estimated as the novelty of a state or state-action pair using visitation counting. 
Exact counting, which is generally intractable in large state-action spaces, can be replaced by Hash-based counting~\citep{hash-exploration}, or methods that focus on direct estimation of the novelty of state-action pairs, such Random Network Distillation~\citep[RND, ][]{rnd}.

In contrast, epistemic uncertainty in a \textit{value} prediction $\V \, [\hat V(s)]$ that is trained with TD-based targets, requires reasoning about uncertainty in the value-targets, as well as whether $\hat V$ was trained on $s$ at all.
We follow the popular approach by \cite{strens2000bayesian} of defining epistemic uncertainty in the value function as the variance of a Bayesian posterior of the Q-values of a policy conditioned on the data the agent has collected, as follows:
\begin{align}
    \label{eq:q_variance_definition}
    \V\, [\hat Q^\pi(s,a)] = \V\, \Big[\hat R(s,a) + \gamma {\textstyle\sum\limits_{s',a'}} \pi(a'|s')\,\hat P(s'|s, a)\,\hat Q^\pi(s',a')\Big] \,.
\end{align}
Note that the epistemic uncertainty about rewards $\V\,[\hat R]$ and transitions $\V\,[\hat P]$ are induced by the data $\mathcal D$ the value function has been trained on, not by the agent's knowledge of the environment. Values predicted by learned value functions can therefore be epistemically uncertain, even if a perfect model of the environment is known to the agent.
In their work on the Uncertainty Bellman Equation (UBE), \citet{o2018uncertainty} propose to approximate an upper bound $ u^\pi(s_t, a_t)  \geq \V\, [\hat Q^\pi(s_t,a_t)]$.
$ u^\pi(s_t, a_t) $ can be learned with (possibly $n$-step) TD targets in a similar manner to value learning from {\em local} uncertainties $\eta(s,a)$:
\begin{equation}
\label{eq:ube_def}
    u^\pi(s_t, a_t) 
    := \eta(s_t, a_t) + 
        \gamma^{2} {\textstyle \sum_{a'}} \pi(a'|s_{t+1})\, u^\pi(s_{t+1}, a')
    \leq \eta(s_t, a_t) + 
        \gamma^{2}\max_{a'} u^\pi(s_{t+1}, a').
\end{equation}
The local uncertainty $\eta$ can be derived from $\V\,[\hat R]$ and $\V\,[\hat P]$. 
Note that the above inequality yields an upper bound on the epistemic uncertainty of {\em any} policy $\pi$ with training data $\mathcal D$.

\subsection{Deep Exploration with Upper Confidence Bounds}
\label{background:UCB_exploration}
A popular approach to harness epistemic uncertainty for deep exploration is that of optimism in the face of uncertainty \citep{Lattimore17}, often formalized using an upper confidence bound (UCB) on the true value of the optimal policy \citep[][]{UCBVI, jin2018q}.
Acting greedily with respect to the maximum UCB guarantees exploration of the environment, as long as the UCB tightens with more samples from the environment.
The efficiency of UCB exploration depends on the epistemic uncertainty estimator.
In environments with small state-action spaces, epistemic uncertainty estimators that predict maximum uncertainty for unvisited states can be used to guarantee exploration of every state-action pair. 
To explore continuous or large state-action domains in practical numbers of samples, using an UCB requires uncertainty estimators that estimate the epistemic uncertainty in unvisited states $s'$ based on a similarity metric between $s'$ and visited states $s \in \mathcal{D}$ \citep{jin2023provably}, such as imposed by RND, for example.

\section{Deep Exploration with Epistemic MCTS}
\label{contributions}
To find an optimal policy, an RL agent must explore the environment until all necessary information is gathered.
When the agent uses search for action selection in the environment, it can search for exploration, exploitation, or a trade-off between the two.
When a learned model $\hat m$ is used in search, the agent faces a problem: the values $q^i_{\hat m}$ in the search tree converge to $ Q_{\hat m}^{*_{\hat m}} $, the $Q$ value \textit{in the learned model} of the policy that is optimal \textit{in that model} $\pi^*_{\hat m}$.
In areas where the learned model is inaccurate, the search may lead to arbitrarily bad actions with respect to expected return in the true environment.
From the perspective of exploration on the other hand, this presents an opportunity: by estimating the epistemic uncertainty the agent can identify the areas where the model is uncertain due to insufficient interactions, and use the uncertainty to direct exploration into these areas.

Our objective is then two-fold: To extend MCTS to estimate and propagate the epistemic uncertainty from the uncertain learned model and harness the epistemic uncertainty in the search to achieve deep exploration of the environment.
To achieve this, we take the following steps:
(i) Formulate the learned model $\hat m$ as a random variable and use it to construct a UCB on $ Q^* $ (Section \ref{contributions_formulating}). 
(ii) Propose search policies to track the maximum UCB exploration objective (Section \ref{contributions_planning_for_exploration}).
(iii) Propagate the epistemic uncertainty through search, such that the epistemic uncertainty $\V\,[q_{\hat m}^i(s,a)]$ in nodes' value predictions $q_{\hat m}^i(s,a)$ can be estimated (Section \ref{contributions_propagating}).
Search with learned transition models introduces additional challenges for uncertainty propagation, which we address in Section~\ref{sec:challenges}.

\subsection{Search with a Learned Reward Model}
\label{contributions_formulating}
For simplicity, we begin by formulating the problem of search with learned models only in terms of a learned reward model and later extend the setup to learned value and transition models.
Consider the following setting: the agent has access to a dataset $ \mathcal{D} = \{(s_i, a_i, r_i, s_{i+1}) \, | \, 0 \leq i < N \}$ of transitions and rewards, 
as well as to the true (possibly stochastic) transition model $P$.
The reward function is bounded with: $ \textstyle {\max_{s,a \in \mathcal{S} \times A} |\mathcal{R}(s,a)| \leq r_{max}} $. 
We define the uncertain model as a random variable $\hat M = (P, \hat R)$.
The uncertain mean-reward function is defined as a random variable $\hat R$ in the standard manner for defining an epistemically uncertain model (see Section~\ref{background_uncertainty}), such that
$ \E[\hat R(s,a)] = r(s,a) $ the mean reward, and
$ 
\hat R(s, a) = \frac{1}{|(s, a, \cdot, \cdot) \in D|}\sum_{(s, a, r, \cdot) \in D} r \approx r(s,a)
$, as the empirical mean of observed rewards.
The epistemic uncertainty in reward prediction is defined as the variance $ \V\,[\hat R(s, a)]$.
We note that $ \hat R(s, a) $ is a bounded random variable over the interval 
$ [-r_{max}, r_{max}] $.
To facilitate constructing an upper confidence bound on $Q^*$ we define $ \V\,[\hat R(s, a)] := r_{max}^2, \, \forall (s,a) \notin \mathcal{D} $. That is, for unobserved transitions, the maximum variance possible for a bounded random variable \citep{popoviciu1935equations}.
The value with respect to the Markov chain induced by the random variable $ \hat M $ is then itself a random variable:
\begin{align}
 Q^\pi_{\hat M}(s, a) 
 := 
 \E_{\pi, P} \Big[
 \sum_{i=0}^{\infty}\gamma^i r_i \,
 \Big| \, a_i \sim \pi(s_i), s_{i+1} \sim P (s_{i},a_{i}), r_i = \hat R(s_i,a_i), s_0 = s, a_0 = a
 \Big], 
 \label{epistemic_variance_value_def}
\end{align}
where the expectation is with respect to $\pi$ and $P$ and not $\hat R$.
Using this definition for the value in the model $ Q^\pi_{\hat M} $ and its variance $ \V \, [Q^\pi_{\hat M}] $ we can construct an upper confidence bound on $Q^*$ for deep exploration.
\begin{theorem}
\label{thm:UCB}
For $\hat M$, $\mathcal{M}$, $Q^*$, $ Q^\pi_{\hat M} $ defined as above and $\delta \in (0, 1]$:
\begin{align}
 \label{eq:final_ucb}
 P \bigg (
 Q^*(s,a)
 \leq 
 \max_{\pi} \Big ( Q_{\hat M}^\pi(s,a) + \smallfrac{1}{\sqrt{\delta}}\sqrt{\V\,[Q_{\hat M}^\pi(s,a)] }\Big)
 \bigg ) 
 \,\,\geq\,\, 1 - \delta \,, \quad \forall (s,a) \in \mathcal S \times \mathcal A.
\end{align}
\end{theorem}
The proof is provided in Appendix \ref{proof_UCB} and relies on the linearity of $\textstyle{Q_{\hat M}^\pi}$ in the reward function $\hat R$.
An extended UCB that is maintained in the presence of an uncertain transition model $\hat P$ is constructed in Appendix \ref{EMCTS_learned_f}.
The UCB can be tracked by MCTS using epistemic search policies, which we propose next.

\vspace{-1mm}
\subsection{Planning for Exploration with Epistemic Search}
\label{contributions_planning_for_exploration}
To harness search for deep exploration through the approximation of the UCB in Equation~\ref{eq:final_ucb}, we propose Epistemic P/UCT (EP/UCT, changes are marked in blue):
\begin{align} 
 \label{eq:EUCT}
 a_{\text{\textcolor{blue}{E}UCT}} 
 \;=\; 
 \argmax_{a \in A}
 \textcolor{blue}{q^\beta_{\hat M}(s_k, a)} +
 C_{\text{UCT}} {\textstyle\sqrt{\frac{2 \log (\sum_{a'}\! N(s_k, a'))}{N(s_k, a)}}},
 \\
 \label{eq:EPUCT}
 a_{\text{\textcolor{blue}{E}PUCT}} 
 \;=\; 
 \argmax_{a \in A}
 \textcolor{blue}{q^\beta_{\hat M}(s_k, a)} +
 \pi(a | s_k) \, C_{\text{PUCT}} {\textstyle\frac{ \sqrt{\sum_{a'}\! N(s_k, a')}}{1 + N(s_k, a)}},
\end{align}
where we define:
\begin{align} 
 \textcolor{blue}{q^\beta_{\hat M}(s_k, a)}
 :=
 \underbrace{\textstyle \frac{1}{N(s_k,a)}\sum_{i=0}^{N(s_k,a)} \nu_{\hat M}^i(s_k, a)}_{:= q_{\hat M}(s_k, a)}
 +
 \textcolor{blue}{\beta \underbrace{\textstyle \frac{1}{N(s_k,a)}\sum_{i=0}^{N(s_k,a)} \sqrt{\V\,[\nu_{\hat M}^i(s_k, a)]}}_{:= \sigma_{q_{\hat M}}(s_k, a) \geq \sqrt{\V \,[q_{\hat M}(s_k, a)]} }}. 
 \label{eq:opened_EUCT}
\end{align}
The hyper-parameter $\beta \geq 0$ can be tuned per task, or chosen to guarantee an upper confidence bound with specific confidence $1 - \delta$, and 
$\textstyle{\nu_{\hat M}^i(s_k, a)}$ is as defined in Equation \ref{eq:backup} for a specific model $\textstyle{\hat M}$.
We suppress the dependence on planning step $i$ in the notation of $ \textstyle{q^\beta_{\hat M}(s_k, a)} $ for simplicity.
To maintain the property of PUCT, which assumes $q_{\hat M}$ is between 0 and 1 \citep[][]{rosin2011multi}, one can use the Q normalization approach proposed by \cite{MuZero} to normalize the $q^\beta$ scores.
More modern variations of PUCT such as those proposed by \cite{gumbelmuzero} can be used by replacing $ q_{\hat M} $ with $ q^\beta_{\hat M} $ in the search objective.
We note that in order to properly harness search policies that use a prior-policy for deep exploration, such as EPUCT, the agent must learn an exploration-prior-policy $\pi_e$ that is trained with an exploratory objective.
Otherwise, the prior policy may direct the search away from the exploratory planning objective.
To enable EP/UCT, in the following section we propose methodology to estimate the epistemic uncertainty in backups and upper bound $ \V \,[q_{\hat M}(s_k, a)] $.

\subsection{Propagating Epistemic Uncertainty in Search}
\label{contributions_propagating}
To estimate $ \V \,[q_{\hat M}(s_k, a)] $ we need to: (i) extend the problem setup to account for the learned value model, (ii) compute or upper bound the epistemic uncertainty in one backup $\V\,[\nu_{\hat M}^i(s_k,a)]$ and (iii) compute or upper bound $\V\,[q_{\hat M}(s_k,a)]$ using $\V\,[\nu_{\hat M}^i(s_k,a)]$.

\textbf{Search with a Learned Value Model}
A learned value model $\hat V$ introduces an additional source of epistemic uncertainty into the search, $ \V\,[\hat V(s)] $.
In this case, for the UCB to hold it is important that the models $\hat M = (f, \hat R, \hat V)$ are trained on the same data, which is the popular choice in practice.
This is sufficient to analytically maintain that $ \V\,[\hat V(s)] \geq \V\,[V_{\hat M}^\pi(s)] $ (see Equation \ref{eq:q_variance_definition}).
To maintain this property in practice, we choose the UBE \citep{o2018uncertainty} estimator $\hat u$ (see Section \ref{background_uncertainty} and Equation \ref{eq:ube_def}) which upper bounds $ \V\,[\hat V(s)] $ using $ \V[\hat R(s,a)] $. 
To account for the possibility that $\hat u$ is unreliable outside of the training set 
we use the novelty of $ (s,a) $ to upper bound the uncertainty predicted by $\hat u$ (see Appendix \ref{appendix:ube_predictions} and Equation \ref{eq:ube_max}).
When the reward model is not learned, such as in AZ, the value model still has epistemic uncertainty as it is trained from interactions and $\V\,[\hat V(s)]$ coincides with the definition of value uncertainty in model-free literature (Equation \ref{eq:q_variance_definition}).
This definition accounts for uncertainty in transitions that are unobserved in the environment, regardless of whether the transitions are known to the planning model, capturing the uncertainty in a value model that is trained from interactions.
To estimate $ \V\,[\hat R(s, a)] $ RND and (pseudo-)counting methods can be used (see Section \ref{background_uncertainty} for more detail).

\textbf{Epistemic Uncertainty of the Backup}
The epistemic uncertainty in one backup step $ \V \, [\nu_{\hat M}^i(s_k, a)] $ starting at node $s_k$ and choosing action $a$ can be formulated as follows:
\begin{align}
 \V \,[\nu_{\hat M}^i(s_k,a)] 
  = 
  \sum_{j = 0}^{T-k-1}\gamma^{2j} 
  \V \,[\hat R(s^i_{k+j}, a^i_{k+j})] + 
  \gamma^{2(T-k)}
  \V \,[\hat V(s^i_{T})],
  \label{eq:return_variance_learned_reward}
\end{align}
under the assumption that $ \hat R(s_{k+i}, a) $ is independent from $ \hat R(s_{k+j}, a) \,, \forall i \neq j, $ and $ \hat V $ is independent from $ \hat R$ conditional on the data.
We note that this assumption might be reasonable for some estimators (i.e., a table), but is not generally assumed to be true for DNNs.
To avoid this assumption, we can instead upper bound the standard deviation of correlated backups using the sum of standard deviations (see Appendix \ref{derivation_variance_q}).

\textbf{Epistemic Uncertainty of Node Values}
Since A/MZ use the same model $\hat M$ throughout planning, the returns predicted for different backups cannot be assumed to be de-correlated. 
We propose to upper bound the epistemic uncertainty instead:
\begin{align}
 \V\, [q_{\hat M}(s_k,a)]
 \leq 
 \sigma^2_{q_{\hat M}}(s_k,a) 
 := 
 \bigg ( \frac{1}{N(s_k,a)}
  \sum_{i=0}^{N(s_k,a)} \sqrt{\V\,[\nu^i(s_k,a)]}
 \bigg)^{\!\!2},
 \label{eq:ucb_variance}
\end{align}
where $ N(s_k,a) $ is the number of visitations to action $a$ at node $s_k$.
We provide a complete derivation in Appendix \ref{derivation_variance_q}.
Equation \ref{eq:ucb_variance} completes the Epistemic MCTS (EMCTS) algorithm for learned value and/or reward models.
See Algorithm \ref{alg:MCTS_modifications} for pseudo code of EMCTS with EUCT, where we suppress dependence on the model $\hat M$ for notation simplicity. 
Extensions to MCTS are marked in \textcolor{blue}{blue}.
In AZ, with a true reward model, EMCTS will use $ \V\,[\hat R(s, a)] = 0, \, \forall (s, a) \in \mathcal{S} \times \mathcal{A} $, but still use $\V\,[\hat V(s)] \geq 0$.

To conclude this section we note that while in this work we motivate and later evaluate EMCTS from the perspective of deep exploration, EMCTS is not limited to this use case.
Our method introduces to MCTS-based algorithms such as A/MZ a novel capability to estimate the epistemic uncertainty in the value predictions during and post search, which may be attractive for different purposes.
Noted example are:
(i)~reducing over estimation errors with lower-bound value targets of the form $ q_{\hat M}(s,a) - \beta \sqrt{\V \, [q_{\hat M}(s,a)]} $. This approach is popular in model free AC methods implicitly~\citep{td3,sac} and explicitly~\citep{ciosek2019better}, as well as in model-based RL \citep{zhou2020deep} and is generally very common in offline RL~\citep{kumar2020conservative,ghasemipour2022so}. This makes EMCTS an espeically attractive candidate to enhance A/MZ's Reanalyze for offline-RL or off-policy target generation~\citep{schrittwieser2021online}.
We propose to track the objective $ q_{\hat M}(s,a) - \beta \sqrt{\V \, [q_{\hat M}(s,a)]} $ in search, i.e. to use EP/UCT but with $\beta < 0$.
(ii)~Weighting value and policy losses by the estimate of the epistemic uncertainty in the value which was successful in online as well as offline RL~\citep{lee2021sunrise, u_w_ac}.
The uncertainty of the value of the root of EMCTS can be used for this purpose.

\setlength{\textfloatsep}{2mm} % Remove space after alg.
\begin{algorithm}[tb]
    \caption{EMCTS with EUCT. Requires $f, r, v^\pi$ \textcolor{blue}{and uncertainty estimators 
    $ \V \, [\hat V], \V\,[\hat R]$
    }}
    \label{alg:MCTS_modifications}
    \begin{algorithmic}[1]
        \vspace{.5mm}
        \Function{EMCTS}{state $s$, \textcolor{blue}{$\beta$}}
        \Comment{$\beta=0$ for unmodified MCTS exploitation episodes}
            \While{within computation budget}
                \State SELECT($s$, \textcolor{blue}{$\beta$})
                \Comment{traverses tree from root $s_0 := s$ and adds new leaf}
            \EndWhile
            \vspace{-.5mm}
            \State \Return action $a$ drawn from
            $\pi(a_0|s) = \frac{N(s_0, a)}{\sum_{a'} N(s_0, a')}$
            \Comment{MCTS action selection}
        \EndFunction
        \vspace{.5mm}
        \Function{SELECT}{\text{node} $s_{k}$, \textcolor{blue}{$\beta$}}
            \vspace{-1.5mm}
            \State $ a_{k} \gets \argmax_{a \in A}
            \textcolor{blue}{q^\beta(s_k, a)} +
            C_{\text{UCT}} {\textstyle\sqrt{\frac{2 \log (\sum_{a'}\! N(s_k, a'))}{N(s_k, a)}}} $
            \Comment{Equation \ref{eq:EUCT}}

            \State \textbf{if} {$a_k$ already expanded}
                \textbf{then} SELECT($f(s_k, a_k), \textcolor{blue}{\beta}$)
                \Comment{traverses tree}
            \State \textbf{else} EXPAND($s_k, a_k$)
                \Comment{adds new leaf}
        \EndFunction
        \vspace{1mm}
        \Function{EXPAND}{node $s_{k}$, not yet expanded action $a_{k}$}
            \vspace{.5mm}
            \State $s_{k+1}, \hat V(s_{k+1}), \hat R(s_k, a_k) $ $\gets$ \text{Execute MCTS expansion} 
            \Comment{creates a new leaf $s_{k+1}$}
            \State \textcolor{blue}{Estimate epistemic variance of reward $\V\,[\hat R(s_k, a_k)]$ and store it in node $s_{k+1}$.}
            % \State \textcolor{blue}{
            % $
            % \sigma^2_{\hat R}(s_k, a_k) \gets 
            % \V\,[\hat R(s_k, a_k)]
            % $} \Comment{Estimate reward variance}
            \State \textcolor{blue}{Estimate epistemic variance of value $\V\,[\hat V(s_{k+1})]$ and store it in node $s_{k+1}$.}
            % \State \textcolor{blue}{
            % $
            % \sigma^2_{\hat V}(s_k) \gets 
            % \V\,[\hat V(s_{k+1})]
            % $} 
            % \Comment{Estimate value variance}
            \vspace{.5mm}
            %\State $\sigma_q^2(s_k, a_k) \gets \V\,[R_k] + \gamma^2 \V\,[V_{k+1}]$
            %\Comment{node attribute of $s_{k+1}$, Equation \ref{value_variance_node}}
            %\vspace{.5mm}
            %\If{$k > 0$}
            \vspace{-.25mm}
                \State BACKUP($s_{k+1}, \hat V(s_{k+1}), \textcolor{blue}{\V\,[\hat V(s_{k+1})]}$)
                \Comment{updates the tree values \& value variances}
            %\EndIf
        \EndFunction
        \vspace{1mm}
        \Function{BACKUP}{node $s_{k+1}$, ret. $\nu(s_{k+1}, a_{k+1})$, \textcolor{blue}{ret. unc. $\V\,[\nu(s_{k+1}, a_{k+1})]$}}
            \vspace{.5mm}
            \State $s_k, a_k, \nu(s_k, a_k) \gets$ \text{Execute MCTS backup step} 
            \Comment{updates $q(s_k, a_k), \, N(s_k, a_k)$\!\!\!}
            \vspace{1mm}
            \State \textcolor{blue}{$\V\,[\nu(s_k, a_k)] \gets \V\,[\hat R(s_k, a_k)] + \gamma^2 \V\,[\nu(s_{k+1}, a_{k+1})]$}
            \Comment{Equation \ref{eq:return_variance_learned_reward}}
            
            % \vspace{-.25mm}
            \State \textcolor{blue}{$\sigma_{q}(s_k, a_k) \gets 
            \sigma_q(s_k, a_k) + \smallfrac{\sqrt{\V\,[\nu(s_k, a_k)]} - \sigma_q(s_k, a_k)}{N(s_k, a_k)}$}
            \Comment{Equation \ref{eq:ucb_variance}}
            
            \vspace{0mm}
            \State \textbf{if} {$k > 0$} \textbf{then}
                BACKUP($s_k, \nu(s_k, a_k), \textcolor{blue}{\V\,[\nu(s_k, a_k)]}$)
                % \Comment{updates the tree values \& variances}
        \EndFunction
    \end{algorithmic}
\end{algorithm}
\subsection{Search with a Learned Transition Model}
\label{sec:challenges}
Learned transition models introduce several challenges from the perspective of EMCTS: 
(i) estimating epistemic uncertainty in the possibly-abstracted planning space, 
(ii) propagating the uncertainty forward during search as future transitions become less certain, in addition to backwards, and (iii) propagating it in such a way that maintains the UCB constructed in Theorem \ref{thm:UCB}.
Multiple methods to overcome (i) have been successful in previous works, see \citet{henaff2019explicit, sekar2020planning}.
In Appendix \ref{EMCTS_learned_f} we propose an approach to overcome (ii) and (iii) as well as discuss the challenges in more detail. 
In practice however, in Section \ref{results} we include experiments where a MZ agent fitted with a reliable uncertainty estimator for $\V\,[\hat{R}(s,a)]$ successfully demonstrates deep exploration without accounting for (ii) and (iii).

\vspace{-1mm}
\section{Related Work}
\label{Related_work}
Search for exploration was used successfully in a number of previous works, see 
\cite{UCRL2}, 
\cite{yi2011planning},
\cite{hester2012intrinsically},
\cite{shyam2019model},
\cite{henaff2019explicit},
\cite{sekar2020planning},
\cite{lambert2022challenges} and 
\cite{luis2023model}.
We add to this line of work EMCTS: 
designed for MCTS (and planning trees in general), practical and based in theory.
\cite{bayes_mcts} develop a Bayesian approach for aleatoric uncertainty propagation in MCTS.
POMCP \citep{POMCP}, POMCPOW \citep{POMCPOW} and BOMCP \citep{BOMCP} extend MCTS to POMDPs with a probabilistic Bayesian belief state at the nodes using a probabilistic model, while Stochastic MuZero \citep{antonoglou2021planning} extends MuZero to the stochastic transitions setting by replacing the deterministic transition function with a Vector Quantised Variational AutoEncoder \citep{van2017neural}. 
In these works, epistemic uncertainty is not distinguished or used for exploration.
Latent disagreement ensembles \citep{lakshminarayanan2017simple, ramesh2022exploring} offer a popular alternative to RND and counts.
Wasserstein Temporal Difference \citep[WTD,][]{metelli2019propagating} offers an alternative to UBE \citep{o2018uncertainty} for estimating value uncertainty, using Wasserstein Barycenters \citep{agueh2011barycenters} to update a posterior over $Q$ functions in place of a standard Bayesian update.
While UBE was criticized by \cite{janz2019successor} for being insufficient for deep exploration with posterior-sampling based RL \citep[PSRL,][]{osband2013more}, we note that the same shortcomings do not apply when UBE is used for UCB-based exploration.

\vspace{-1mm}
\section{Experiments}
\label{results}
The task of writing code and finding new algorithms, where AZ has recently made world real-world impact \citep{AlphaTensor, AlphaDev}, is natural to formulate using sparse-reward environments where the actions of the agent are operations and reward is received when a correct program is completed.
Such environments represent a hard exploration challenge: the state space is often exponential $|\mathcal A|^L$ in the actions $\mathcal A$ and the maximum length of the program $L$,
and while the number of possible solutions is generally unknown, it is in most cases very small compared to the number of possible sequences of operations.
To evaluate the contribution of EMCTS in such real-world tasks we conduct experiments in the one-instruction Assembly programming language {\sc subleq} (Section~\ref{sec:experiments_subleq}).

To verify that EMCTS demonstrates deep exploration and benefits from search, we conduct experiments in the commonly used hard-exploration benchmark Deep Sea \citep{osband2019behaviour} (Section \ref{sec:experiments_deep_sea}).
Specifically, we are interested in the following: \\
\textbf{RQ I}
\textit{Does EMCTS paired with an epistemic search policy demonstrate deep exploration with both AZ as well as MZ and in the presence of stochastic as well as deterministic rewards?}\\
\textbf{RQ II}
\textit{Is there benefit in search with EMCTS for deep exploration, compared to an otherwise equivalent approach that does not use search, and if so, is the benefit retained in the presence of the learned transition model of MZ?} \\
We also conduct an ablation study on the exploration parameter $\beta$ to verify that the agent can learn stably from the possibly very-off-policy exploratory data provided by deep exploration.

\subsection{{\sc subleq} Experiments}
\label{sec:experiments_subleq}
\begin{figure}[h]
  \vspace{-1mm}
  \centering
  \includegraphics[width=1.0\linewidth]{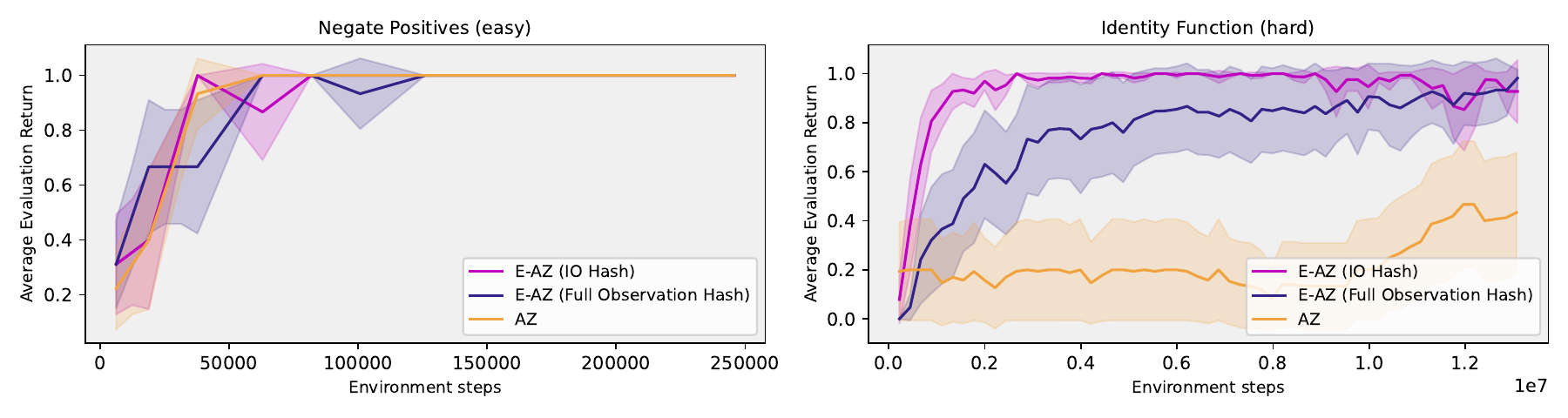}
  \vspace{-7mm}
  \caption{
  Sample efficiency. Left: the easy \textsc{subleq} \textit{Negate Positives} task. Right: the harder \textit{Identity Function} task. Mean of 15 seeds, two standard errors.
  }
  \label{fig:results_subleq}
  \vspace{-1mm}
\end{figure}

{\sc subleq} is a Turing-complete (excepting for finite memory) one-instruction programming language \citep{subleq}.
Because there is only one instruction, writing code in {\sc subleq} summarizes to writing a sequence of memory addresses.
We model the action space with $\mathcal{A} = \{0, \dots, N-1\}$, for $N$ the size of the memory.
The agent is rewarded with $1.0$ when the sequence of addresses specifies a program that solves the task, evaluated on a set of test cases.
The reward is otherwise zero.
The observation space constitutes of an example input and corresponding desired output, the program written so far, and the state of the input and output after executing the program on the example input.
We limit the memory size to 16 and the maximum program length to $10$ resulting in a state space of size $ \leq 16^{10} \approx 1 $ \textit{trillion} unique states.
We present results on two tasks: An easier task of outputting the negated input for positive inputs (Negate Positives), and a harder task of implementing the Identity Function.
An implementation of Negate Positives in {\sc subleq} of length 2 is known, which means finding a solution without prior knowledge requires searching on the order of $16^2$ unique states.
To the Identity Function an implementation of length $ 6 $ is known which suggests searching on the order of $16^6 \approx 16 $ million unique states.
We provide a more detailed introduction to {\sc subleq}, a description of the environment and the tasks in Appendix~\ref{app:subleq_explanation}.
The results are presented in Figure~\ref{fig:results_subleq}.

We compare two variations of EMCTS with AZ (\textbf{E-AZ}) to the \textbf{AZ} baseline.
Both variations of E-AZ are able to solve the harder task in a much smaller number of samples than AZ.
To estimate $\V\,[\hat R(\cdot,\cdot, s')]$ one variation of E-AZ uses a hash based visit count that takes as input the complete state~$s'$, allowing the agent to, in principle, avoid searching the same state twice.
The other E-AZ variation hashes only part of the state: the example input-output, before and after execution (IO hash), directing the agent to search actions that have an effect on the output of the program.
As expected from a UCB-based method, E-AZ benefits from the more-appropriate uncertainty estimator and finds a correct program much earlier with the IO hash.

All agents use \cite{gumbelmuzero}'s approach that combines Gumbel noise with Sequential Halving at the root and a modern variation of PUCT at non-root nodes.
For E-AZ variations, in online search (when selecting actions in the environment) the tree search policies use $q^\beta_{\hat M}$ (\ref{eq:opened_EUCT}) in place of $ q_{\hat M} $ and an exploration-prior-policy $\pi_e$ (See Section~\ref{contributions_planning_for_exploration}).
To select actions during evaluation, as well as when generating targets with Reanalyze, the search uses EMCTS to propagate the uncertainty, but uses the exploitation prior policy $\pi$ and $ q_{\hat M} $ in the search objective, in the standard manner of MCTS search.
The policy $\pi_e(s)$ is trained with cross entropy loss to fit the softmax across actions over $q^\beta(s,\cdot)$ at the root.

\subsection{Deep Sea Experiments}
\label{sec:experiments_deep_sea}
\begin{figure}[h]
  \vspace{-1mm}
  \centering
  \includegraphics[width=1.0\linewidth]{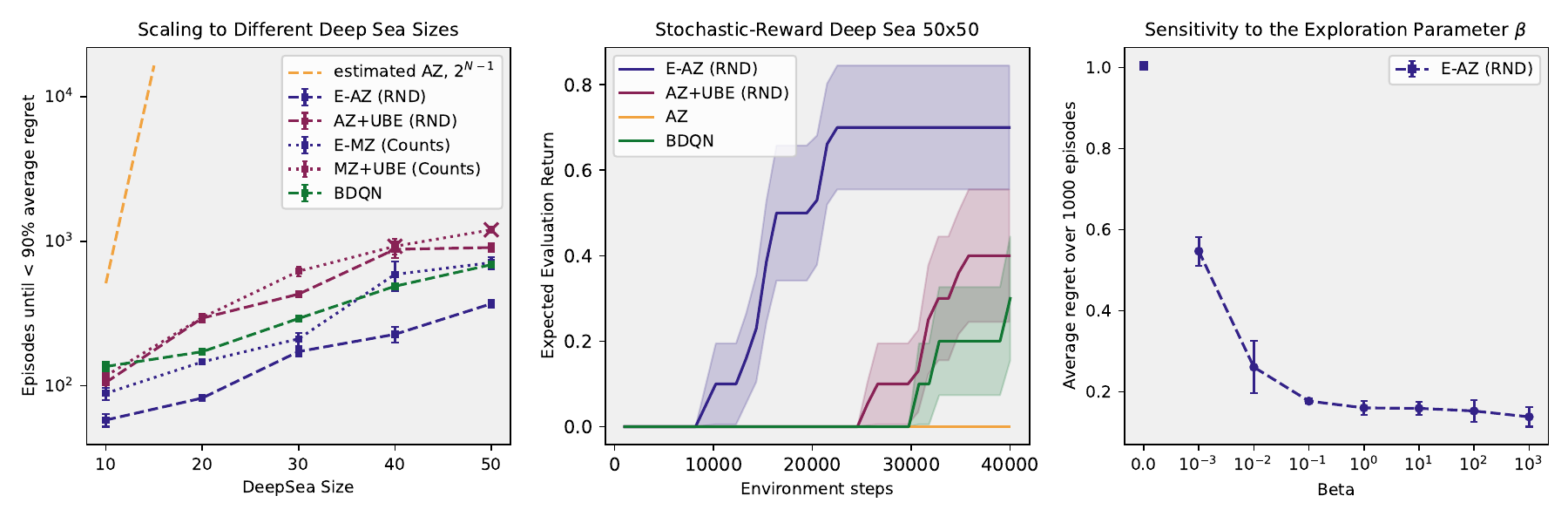}
  \vspace{-7mm}
  \caption{
  Left: Scaling to growing Deep Sea sizes, 5 seeds per point.
  Only 2 seeds of MZ+UBE were able to solve size 40, and none size 50 within the training budget, both marked with an X.
  Middle: Stochastic-reward Deep Sea 50x50, 10 seeds. 
  Right: The effect of the exploration parameter in Deep Sea 30x30, 3 seeds per point.
  Mean and standard error.
  }
  \label{fig:results_beta_and_scaling}
  \vspace{-2mm}
\end{figure}

The Deep Sea environment is structured as a grid, where at each time step the agent chooses between two actions, goes right or left and one row down.
There is a reward $r_{goal} = 1.0$ at the bottom right corner, the agent starts at the top left corner and thus the probability of randomly finding the unique optimal trajectory decays exponentially with the size of the grid. 
Every transition along the optimal trajectory receives a negative reward that is negligible in comparison to the goal reward, but is otherwise the only reward the agent sees, actively discouraging exploration along the optimal path. 
The action mappings are randomized such that the effect of the same action is different in different states, preventing the agent from generalizing across actions.
To evaluate EMCTS in the presence of deterministic transitions and stochastic rewards that more generally align with the assumptions in Section~\ref{contributions}, we include a custom stochastic-reward variation of Deep Sea where the goal reward $ r_{goal} \sim \mathcal N(1,1) $, and a reward $ r_{mislead} \sim \mathcal N(0, 1) $ is given at the bottom left corner. 
The agent must explore both transitions a sufficient number of times to correctly identify the larger mean reward.
Results are presented in Figure~\ref{fig:results_beta_and_scaling}.

The left subplot shows sample complexity measured as number of interactions until 90\% regret is reached against the size of the Deep Sea environment to demonstrate that the scaling is sub-exponential as expected from deep-exploration methods.
The middle subplot presents expected return in evaluation episodes in the stochastic-reward variation of Deep Sea.
In the right subplot, the sensitivity to the exploration parameter $\beta$ is studied and the ability of the agent to learn stably from off-policy data.
In Appendix \ref{appendix_additional_results} we include additional results, including a visualization of the uncertainty estimated by EMCTS.

To answer the research questions the following agents based in A/MZ are compared:
(i) \textbf{E-A/MZ} (purple, dashed for AZ and dotted for MZ), our method. The agents use EMCTS with counts/RND to estimate reward and transition uncertainty and UBE to estimate value uncertainty, search with EUCT and act with the action with the most visitations at the root.
(ii)~Baseline~\textbf{AZ} (orange) explores by sampling actions proportionally to visitations at the MCTS root. 
AZ is included for reference, to demonstrate that indeed Deep Sea cannot be solved in reasonable time with random exploration, and is not able to solve any of the environments in the alloted training steps, with the exception of one seed in the smallest environment size, and that only because the initial replay buffer by chance already contained a trajectory that reached the goal.
We include the exponentially-scaling performance expected of random-exploration based methods such as AZ, for reference.
(iii) Last, \textbf{A/MZ+UBE} (red) acts with a similar UCB-exploration objective to that of EP/UCT, but does not search with the uncertainty (see Appendix~\ref{app:only_ube}).
AZ-based agents are given access to the true transition model. 
The value model is learned, as well as the reward model to investigate the behavior under the most general setting considered in Section \ref{contributions_propagating}.
All A/MZ agents are trained with targets generated by Reanalyze. 
For reference performance on Deep Sea, we include Bootstrapped DQN \citep[\textbf{BDQN, }][]{osband2018randomized}, a popular model-free, non-search based deep exploration approach that relies on an ensemble to drive deep exploration directly.
For full implementation details see Appendix~\ref{appendix_implementation}.

As expected for deep exploration methods, agents that are informed with respect to epistemic uncertainty (E-A/MZ, A/MZ+UBE and BDQN) demonstrate sample efficiency that scales sub-exponentially with environment size (Figure~\ref{fig:results_beta_and_scaling}, left).
In addition, E-AZ is able to solve Deep Sea in the presence of stochastic rewards (Figure~\ref{fig:results_beta_and_scaling}, middle).
This answers \textbf{RQ I}: EMCTS successfully demonstrates deep exploration in the presence of learned value, reward and transition models as well as both deterministic and stochastic rewards.
E-A/MZ (Figure~\ref{fig:results_beta_and_scaling}, left and middle plots, purple) demonstrate significant improvement in sample efficiency over the equivalent agents that do not use search (red), in deterministic and stochastic reward environments and even with the learned transition model of MZ (dotted line).
This answers \textbf{RQ II}: EMCTS demonstrates benefits from search for uncertainty estimation and exploration which is retained in the presence of MZ's learned transition dynamics' model.
The low regret in evaluation with exponentially increasing values of $\beta$ (Figure~\ref{fig:results_beta_and_scaling}, right) demonstrates that the agent can stably learn the optimal policy even in the presence of off-policy exploratory data.

\section{Conclusions}
\label{Conclusions}
In this work we present EMCTS, a novel, practical and theoretically motivated method to incorporate the epistemic uncertainty from learned models into MCTS, as well as harness the search for deep exploration.
AZ paired with EMCTS (E-AZ) achieves significantly higher sample efficiency in the sparse-reward, challenging task of programming in the Assembly language {\sc subleq}, compared to baseline AZ.
In the popular hard-exploration benchmark Deep Sea, E-A/MZ demonstrate deep exploration by solving variations of the task  which cannot be solved by baseline A/MZ at all in a reasonable amount of samples.
EMCTS' search demonstrates significantly improved epistemic uncertainty estimation through more sample efficient exploration over an otherwise equivalent method that does not use search.
With EMCTS, A/MZ are much better equipped for sparse-reward and hard-exploration environments, which come up in realistic settings such as algorithm design, where AZ has already made significant advances.
By making A/MZ uncertainty aware, EMCTS is also promising for settings that require improved reliability in the face of the unknown such as offline RL and off-policy target generation.

\section*{Acknowledgements}
\label{Acknowledgements}
We would like to thank 
Marco Loog, 
Frank van der Meulen, 
Itamar Sher, 
Moritz Zanger, 
Pascal van der Vaart, 	
Joery de Vries 
\& 
Jinke He 
for many fruitful discussions and helpful comments.
We acknowledge the use of computational resources of the DelftBlue supercomputer, provided by Delft High Performance Computing Centre (https://www.tudelft.nl/dhpc) as well as the DAIC cluster.
This work was partially supported by the EU Horizon 2020 programme under grant number 964505 (Epistemic AI),
and partially funded by the Dutch Research Council (NWO) project {\em Reliable Out-of-Distribution Generalization in Deep Reinforcement Learning} with project number OCENW.M.21.234.

\bibliographystyle{iclr2025_conference}
\bibliography{bib}

\appendix

\section{Proofs and Derivations}
\label{appendix_proofs}

\subsection{Proof of Theorem \ref{thm:UCB}}
\label{proof_UCB}
For clarity, let us again define $\pi^*$ as the policy that is optimal in the true environment $ \pi^* = \argmax_\pi Q_{m}^\pi $, and $ \pi_{\hat M}^* $ the policy that is optimal in a specific model $\hat M$, that is $ \pi_{\hat M}^* =  \argmax_\pi Q_{\hat M}^\pi $.
By the definition of the model, we have $ \E[\hat M] = m $, where $m$ is the true model of the environment.
Therefore $ Q^{*} =: Q^{\pi^*}_{m} = Q^{\pi^*}_{\E[\hat M]} =\E_{\hat M} [Q_{\hat M}^{\pi^*} ] $, by linearity of the value in the reward function.
By Chebyshev's inequality:
\begin{align}
    P \Big (\E_{\hat M} [Q_{\hat M}^{\pi^*} ]
    \leq 
    Q_{\hat M}^{\pi^*} + \frac{1}{\sqrt{\delta}}\sqrt{\V \,[Q_{\hat M}^{\pi^*}]}\Big ) 
    \geq 1 - \delta.
\end{align}
The right-hand side term $ Q_{\hat M}^{\pi^*} + \frac{1}{\sqrt{\delta}}\sqrt{\V \,[Q_{\hat M}^{\pi^*}]} $ is hard to estimate among other reasons because the optimal policy is not known.
But it can be upper bounded as follows:
$$ Q_{\hat M}^{\pi^*} + \frac{1}{\sqrt{\delta}}\sqrt{\V\,[Q_{\hat M}^{\pi^*}]}
    \leq 
    \max_{\pi} Q_{\hat M}^\pi + \frac{1}{\sqrt{\delta}}\sqrt{\V\,[Q_{\hat M}^\pi]}, $$
which holds because the right-hand side is the maximum possible policy in a set that contains $ \pi^* $.
    
Putting the entire set of inequalities together, we arrive at:
\begin{align}
    P \Big (
    Q^* 
    =
    \E_{\hat M} [Q_{\hat M}^{\pi^*} ]
    \leq 
    Q_{\hat M}^{\pi^*} + \smallfrac{1}{\sqrt{\delta}}\sqrt{\V\,[Q_{\hat M}^{\pi^*}]}
    \leq 
    \max_{\pi} Q_{\hat M}^\pi + \smallfrac{1}{\sqrt{\delta}}\sqrt{\V\,[Q_{\hat M}^\pi]}
    \Big ) 
    \geq 1 - \delta,
    \\
    \text{and thus:} \quad P \Big (
    Q^*
    \leq 
    \max_{\pi} Q_{\hat M}^\pi + \smallfrac{1}{\sqrt{\delta}}\sqrt{\V\,[Q_{\hat M}^\pi]}
    \Big ) 
    \geq 1 - \delta,
\end{align}
QED Theorem \ref{thm:UCB}.

\subsection{Derivation of the Upper Bound on $ \V\,[q_{\hat M}(s_k,a)]$}
\label{derivation_variance_q}
By definition, 
\begin{align}
    \V\,[q_{\hat M}(s_k,a)]
    = 
    \smallfrac{1}{n^2}
        \V\,\Big[
            \smallsum{i=1}{n}
            \nu^i(s_k, a)
        \Big]
    =
        \smallfrac{1}{n^2}
        \smallsum{i,j=1}{n}
        \text{Cov}\big(\nu^i(s_k, a), \nu^j(s_k, a)\big) \,.
\end{align}
In A/MZ it is standard to learn one set of deterministic models  $\hat V, \hat R $ and use them throughout planning, and therefore $ \nu^i, \nu^j $ cannot be assumed to be independent $ \forall i \neq j $.
Had they been independent, one would compute the variance of the sum directly with the sum of the variances.
Instead, we use the inequality for covariances with known variances \citep{Hoessjer22}: 
$\text{Cov}[X, Y]~\leq~\sqrt{\V\,[X] \V\,[Y]}$ to upper bound the variance $ \V\,[q_{\hat M}(s_k,a)] $:
\begin{align}
    \V\,[q_{\hat M}(s_k,a)]
    = 
    \smallfrac{1}{n^2}
        \V\,\Big[
            \smallsum{i=1}{n}
            \nu^i(s_k, a)
        \Big]
    =
        \smallfrac{1}{n^2}
        \smallsum{i,j=1}{n}
        \text{Cov}\big(\nu^i(s_k, a), \nu^j(s_k, a)\big)
    \\
    \leq \smallfrac{1}{n^2}
        \smallsum{i,j=1}{n}
        \sqrt{\V\,[\nu^i(s_k, a)]\,
            \V\,[\nu^j(s_k, a)]}
    \;=\; 
    \Big( \smallfrac{1}{n}
            \smallsum{i=1}{n}
            \sqrt{\V\,[\nu^i(s_k, a)]}
        \Big)^{\!2}.
\end{align}
In other words, the standard deviation of the averaged backup $ \sqrt{\V\,[q_{\hat M}(s_k,a)]} $ is upper bounded with the averaged standard deviation across backups $\smallfrac{1}{n}\smallsum{i=1}{n}\sqrt{\V\,[\nu^i(s_k, a)]}$.
The first inequality is due to the inequality for covariances with known variances,
and the following equality is a sum of squares.

\section{EMCTS with learned transition dynamics}
\label{EMCTS_learned_f}
When the transition function $\hat f$ is learned, we must formulate it as a random variable $\hat F$ as well.
We note that in this case $ S' \sim \hat F(S,a) $ is a random variable that depends on the distribution of the previous state $S$ propagated through the uncertain transition function $\hat F$, and similarly the distribution of $S'$ will propagate through the reward and value and future transition predictions $ \hat R(S', a), \hat V_{\hat M}^\pi (S'), \hat F(S', a) $.
In other words, in order to estimate $\V \,[Q^\pi_{\hat M}]$ we need to account for propagation of the variance through the Markov chain, which is an expensive and non-trivial process.
In addition, as the value is non-linear in the transition dynamics $f$, Theorem \ref{thm:UCB} does not apply and Equation \ref{eq:final_ucb} is not an upper bound with a specific probability.
Finally, to extend EMCTS to MZ fully, we must account for the fact that the dynamics model learned by MZ is not operating in the true state space of the environment, but in a value-equivalent abstraction.
To extend EMCTS to the learned dynamics case, we will assume that it is possible to estimate epistemic uncertainty in the abstracted state space $\hat {\mathcal{S}}$ in a meaningful way. 
This problem can be circumvented by driving additional losses through the learned model that incentivize distinguishing between unique states in latent space \citep{henaff2019explicit}, or by learning an auxiliary dynamics model to distinguish between novel and observed starting-states-and-action-sequences, which has been used successfully by \cite{sekar2020planning}. 
We will therefore use the notation of states $s$ or $S$ in the tree, whether they are in the true state space of the environment $\mathcal{S}$ or the abstracted space of MZ $\mathcal{\hat S}$.
In addition, since standard MZ plans with a deterministic transition model, we extend EMCTS to the setting where either the underlying transition dynamics $f$ are deterministic, or the abstracted deterministic transition function is sufficiently meaningful.
For simplicity, let us further assume that the state space $\mathcal{S}$ is continuous, and the starting state distribution $\rho$ is over a finite domain.

To circumvent the problem of the propagation of state uncertainty through the Markov chain, we propose a cheap and simple maximally-optimistic alternative upper-bound approximation for $ \V \,[Q^\pi_{\hat M}] $.
We note that: (i) $\hat F, \hat R $ are trained on the same data and thus when the epistemic variance over one of the two is maximal the other can be expected to be maximal as well. (ii) We have assumed a deterministic true transition function $f$ and thus the variance in $\hat F$ can be modelled as maximal on the unknown and zero on the known.
(iii) If the uncertainty in the prediction of any state $ \V \,[\hat F(S_k,a)] \geq 0 $, we can expect the uncertainty in all future predictions $ S_j $ along this trajectory $ j > k $ to be associated with maximal uncertainty.
We use these observations to formulate the following simple approximation:
\begin{align}
    \V\,[\hat R(S_k, a)] \approx \V\,[\hat R(s_k, a)], 
    \,\,
    \V\,[\hat V_{\hat M}^\pi(S_k)] \approx \V\,[\hat V_{\hat M}^\pi(s_k)], 
    \,\,
    \forall (s,a) \in \mathcal{D}
    \label{eq:mz_variance_observed}
    \\
    \V\,[\hat R(S_j, a)] \approx r^2_{max}, 
    \,\,
    \V\,[\hat V_{\hat M}^\pi(S_j)] \approx (\smallfrac{1}{1-\gamma}r_{max})^2, 
    \,\,
    \text{where} 
    \,\, j \geq k \,\, \text{and} \,\, \forall (s,a) \notin \mathcal{D}
    \label{eq:mz_variance_unobserved}
\end{align}
That is, we propose to ignore any uncertainty in transition along zero-uncertainty trajectories, and with the first uncertain transition, to assume all uncertainties are maximal for all predictions in the rest of the trajectory.
To identify whether $ (s,a) \notin \mathcal{D}$ we can directly use $ \V\,[\hat R(s, a)]$, where $ \V\,[\hat R(s, a)] \approx r^2_{max} $ indicates $ (s,a) \notin \mathcal{D}$.
This will guarantee that (given that the mechanism to identify unobserved transitions is sufficiently reliable) the agent remains sufficiently optimistic and the planning objective of EP/UCT remains an upper bound on $Q^*$.
We note that in practice, while this approach is theoretically sound, the results of the experiments in Section~\ref{sec:experiments_deep_sea}, in the presence of reliable transition-uncertainty, motivate that it is not necessary and the regular setup of EMCTS is sufficient for deep exploration even in the presence of a learned, value-equivalent-abstraction-based transition model, given that $\hat R(s,a)$ is a sufficiently reliable estimator of novelty in the environment.
\section{Additional Experiments}
\label{appendix_additional_results}
\begin{figure}[ht]
  \vspace{-1.5mm}
  \centering
  \includegraphics[width=.85\linewidth]{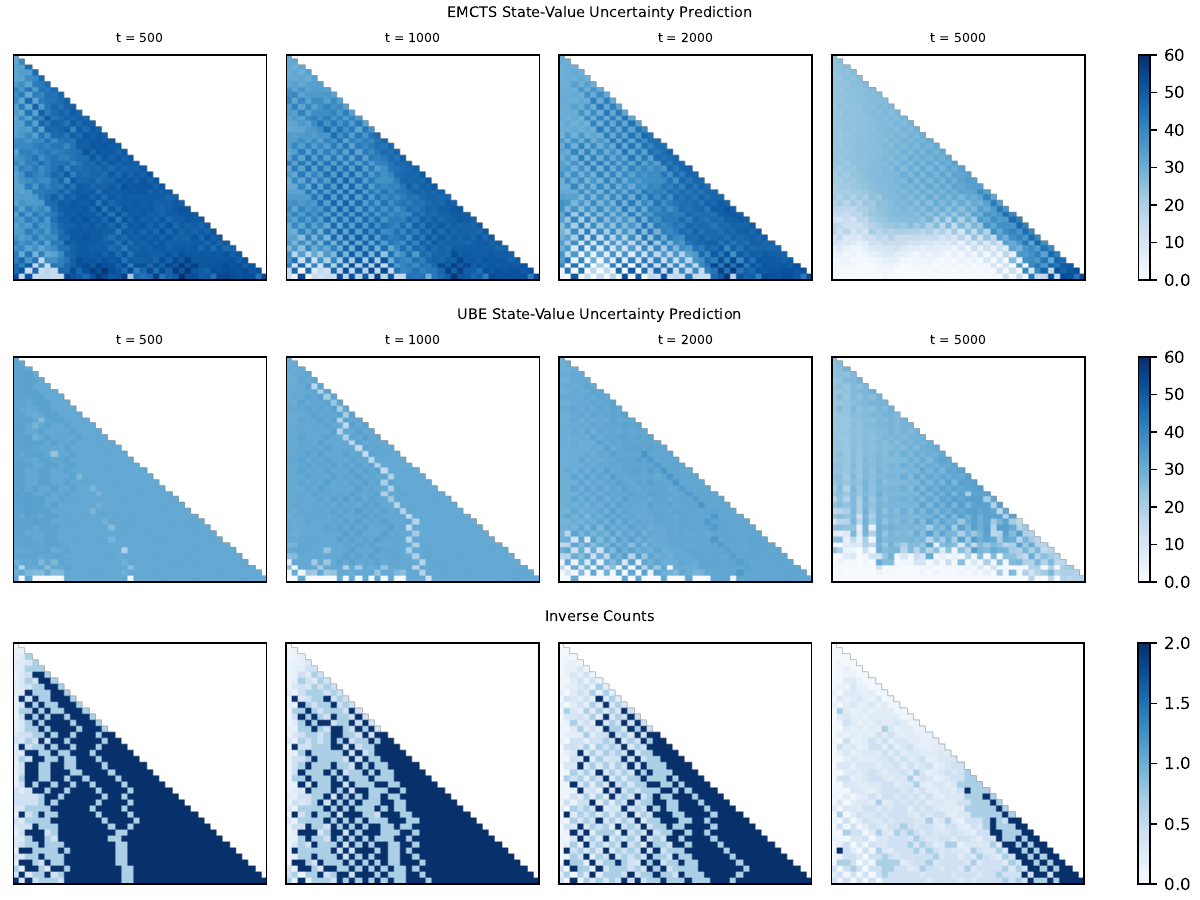}
  \caption{\label{fig:heatmaps}
      Heat maps over states in DeepSea 40 by 40 at different times (columns) during an example training run of EMCTS with an AZ transition model. 
      Upper row: value uncertainty at the EMCTS root node. 
      Middle row: single prediction of UBE at each state.
      Lower row: inverse visitation counts as reliable local uncertainty, where score of 2.0 represents unvisited.
      }
  % \vspace{-2mm}
\end{figure}

We include an example comparison of the uncertainty estimated by EMCTS to that of the uncertainty estimator used by EMCTS, the UBE network head, in Figure~\ref{fig:heatmaps}.
The E-AZ agent used in this experiment uses the true transition model but a learned reward as well as value functions, matching the setup of Section~\ref{contributions_formulating}.
The Deep Sea environment is presented as a grid, where states are the cells including and below the diagonal.
Bold blue in the grids in the bottom row (inverse counts) indicates unvisited states.
By averaging across multiple predictions in search, the uncertainty estimated by EMCTS (top row) is much more varied, more accurately associating less or more uncertainty with states that lead into observed / unobserved trajectories respectively, compared to the UBE predictions (middle row).
Most importantly from the perspective of exploration, EMCTS associates larger uncertainty with states that lead into unobserved directions much more consistently than the single predictions of UBE for each state (for an easily visible example, t$ = 2000$, top of the diagonal).

\begin{figure}[ht]
    \centering
    \includegraphics[width=0.85\linewidth]{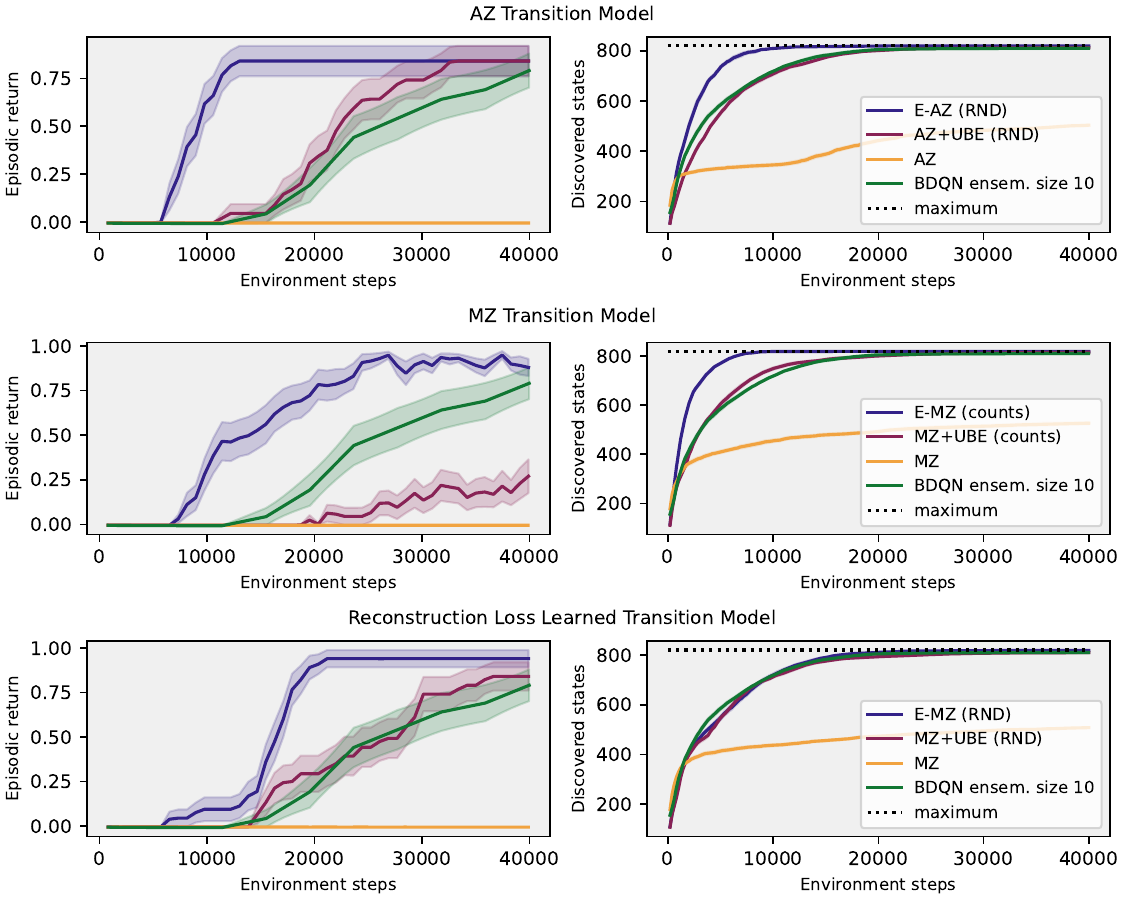}
    \caption{
    Deep Sea 40x40, mean and standard error for 20 seeds.
    Rows: Different transition models.
    Left: episodic return in evaluation vs. environment steps. 
    Right: exploration rate (number of discovered states vs. environment steps).
    }
    \label{fig:results_deep_sea}
    \vspace{-2mm}
\end{figure}

We include an addition detailed evaluation curves for E-A/MZ, A/MZ and A/MZ+UBE in Figure~\ref{fig:results_deep_sea}, as well as rate-of-exploration (number of unique states encountered per interaction with the environment).
An additional variation of MZ is included in this Figure, where instead of value-equivalent abstraction the model is trained with a reconstruction loss to match the true transition function and observation space of the environment, such that $ \hat f(s,a) = \hat s' \approx s'  $.
By training the RND estimator only on true transitions $(s,a)$ and evaluating it online on transitions predicted by the learned transition model $ (\hat s', a'), \, \hat s' = \hat f(\hat s, a) $, this agent is incentivized to estimate the uncertainty over every uncertain transition $(s,a) \notin \mathcal D $ as maximal, implicitly implementing the approach described in Appendix \ref{EMCTS_learned_f}.
For implementation details see Appendix \ref{appendix_dynamics_models}.
In all figures, E-A/MZ outperforms the other agents both in rate of exploration as well as in their ability to learn to reach the goal.
The most interesting behavior is perhaps that of the reconstruction based model, that does not search the environment significantly faster than the other baselines, and yet learns to reach the goal much earlier (bottom row in Figure~\ref{fig:results_deep_sea}).
We hypothesize that due to the search, the uncertainty estimated by the agent is more reliable, resulting in identifying the correct action that leads into novel states, \textit{more times in a row}.
Just visible on the right-hand plot, one can see that indeed the purple curve remains the highest for quite a while, before all others curves match it, despite all curves being very close throughout most of the training.

We include a table evaluating interactions-to-goal on Deep Sea 40x40 for all agents.
The results demonstrate that even when the learned dynamics model is not designed for planning (anchored model, third block, Table \ref{table:first_visit}), EMCTS is able to find the goal much faster. 

\begin{table}[ht]
\centering
\caption{Number of environment steps until the first visitation to the goal transition. 
}
{\small
\begin{tabular}{|l|c|c|c|}
\hline
Novelty Estimator & Exploration & \makecell{Average steps to goal transition for \\ seeds that discovered goal $\pm$ STD} & \makecell{\% seeds that \\ discovered goal}
\\ \hline\hline
\multirow{3}{*}{\begin{tabular}[c]{@{}l@{}} RND\end{tabular}}
                                                                             & E-AZ             & $\boldsymbol{10539} \pm 9006 $            & 94\% of 35 seeds           \\ \cline{2-4} 
                                                                             & AZ+UBE           & $ 22801 \pm 7514$          & 91\% of 35 seeds            \\ \cline{2-4} 
                                                                             & AZ         & -                          & 0\%  of 20 seeds            \\ \hline\hline
\multirow{3}{*}{\begin{tabular}[c]{@{}l@{}} Counts \end{tabular}}
                                                                             & E-MZ             & $\boldsymbol{14339} \pm 6845$           & 100\% of 23 seeds           \\ \cline{2-4} 
                                                                             & MZ+UBE           & $29945 \pm 8113$           & 57\% of 21 seeds            \\ \cline{2-4} 
                                                                             & MZ         & -                          & 0\% of 20 seeds             \\ \hline\hline
\multirow{3}{*}{\begin{tabular}[c]{@{}l@{}}Reconstruction \\ Model (RND)\end{tabular}} 
                                                                             & E-MZ             & $ \boldsymbol{15241} \pm 3236 $         & 95\% of 20 seeds           \\ \cline{2-4} 
                                                                             & MZ+UBE           & $ 22497 \pm 6645 $         & 85\% of 20 seeds            \\ \cline{2-4} 
                                                                             & MZ         & -                          & 0\% of 20 seeds             \\ \hline
\end{tabular}
}
\label{table:first_visit}
\end{table}

\begin{figure}[ht]
    \centering
    \includegraphics[width=1\linewidth]{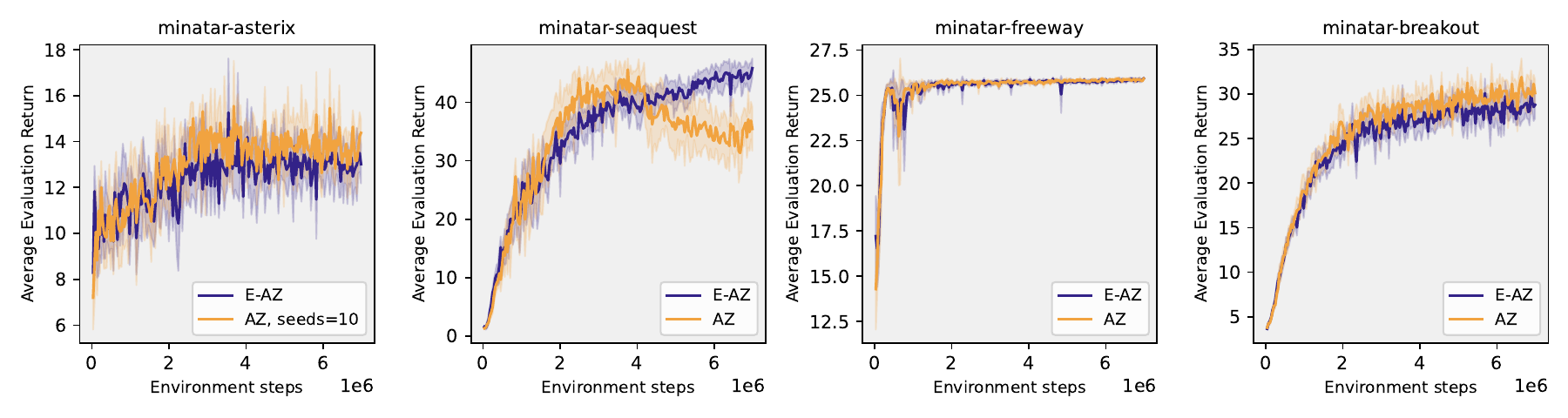}
    \vspace{-4mm}
    \caption{
    Mean and two standard errors for 10 seeds.
    }
    \label{fig:results_minatar}
\end{figure}

Finally, we include a comparison between E-AZ and AZ on MinAtar in Figure \ref{fig:results_minatar}. The uncertainty estimator is the full-Hash used for SUBLEQ, which uses hash based counting to distinguish between any two unique states. 
This demonstrates that even when the uncertainty estimator is unsuited, and even without tuning the exploration parameter $\beta = 0$ (in Figure \ref{fig:results_minatar} $\beta = 1.0$ as in the SUBLEQ experiments), EMCTS can compare well to the baseline.

\def\Dkl{D_\text{KL}}
\newcommand{\Set}[1]{\mathcal{#1}}

\section{Subleq}
\label{app:subleq_explanation}
What follows is a formal characterization of {\sc subleq} as used in our experiments, and example programs for the two tasks we presented.

For {\sc subleq-$N$}, we have a \emph{memory} made up of $N$ \emph{words} $(w_0, w_1, ..., w_{N-1})$ where each word $w_i$ is an integer from $0$ to $N-1$ inclusive (we write $w_i \in [0, N-1]$). We also have an input sequence $(v_0, v_1, ...)$, $v_i \in [0, N-1]$, of variable finite length depending on the particular task to be solved. Tasks correspond to different algorithms we want the agent to implement. Lastly, there is an output buffer $(u_0, u_1, ...)$, $u_i \in [0, N-1]$ which begins empty, but can be extended with output values during execution of a {\sc subleq} program.

We refer to $w_i$ as the word at \emph{address} $i$. Additionally, we give names to some specific addresses: $\texttt{@IN} = N - 3$, $\texttt{@OUT} = N - 2$, $\texttt{@HALT} = N - 1$. The program is stored contiguously in memory starting at address $0$. Execution begins from address $0$ as well, meaning the first instruction to execute is $(w_0, w_1, w_2)$.

When executing some instruction at address $i$, we look at the three words $(w_i, w_{i+1}, w_{i+2})$, and modify the memory by subtracting the value at address $w_{i+1}$ from the word at address $w_i$, i.e. $w_{w_i} \leftarrow w_{w_i} - w_{w_{i+1}}$. We say that this instruction \emph{reads} from $w_i$ and $w_{i+1}$, and \emph{writes} to $w_i$. We also say that reading from $w_i$ \emph{returns} $w_{w_i}$. All operations are done modulo $N$ so that the words remain in the range $[0, N-1]$. If the result before modulo was less or equal to zero, execution continues from address $w_{i+2}$ (we say we \emph{jump} to $w_{i+2}$), meaning the next instruction would be $(w_{w_{i+2}}, w_{w_{i+2}+1}, w_{w_{i+2}+2})$. If the result before modulo was strictly positive, then the next instruction is the next three words, i.e. $(w_{i+3}, w_{i+4}, w_{i+5})$.

Interacting with the addresses \texttt{@IN}, \texttt{@OUT}, and \texttt{@HALT} has different behaviour than normal execution. Reading from the address \texttt{@IN} will instead read the next number from the input sequence. The first read from \texttt{@IN} will return $v_0$, the next read returns $v_1$, etc. Reading from the input sequence when there is no next value will return $0$ instead. Writing to \texttt{@IN} is ignored.

Reading from \texttt{@OUT} always returns a $0$, but writing to \texttt{@OUT} will write to the output buffer. We say that we \emph{output} the value. Each value that we output is added to the output buffer, so after the first output $u_0$, the buffer looks like $(u_0)$, after the second output $u_1$, it is $(u_0, u_1)$, etc.

The output buffer is compared against the desired output for the task to determine whether the task was solved correctly. The first incorrect output terminates the program and results in a failure of the task. If the output buffer ever equals the desired task output, the program terminates and the task is solved successfully. Finally, if the program tries jumping to an address where the instruction would overlap \texttt{@HALT}, such as $(w_{\texttt{@IN}}, w_{\texttt{@OUT}}, w_{\texttt{@HALT}})$, the program terminates.

% Subleq as an RL environment
When using {\sc SUBLEQ-$N$} as a reinforcement learning environment, the agent writes a program one word at a time. It gets to observe the current state of the memory before execution (it is just the program it has written up to that point followed by zeroes), an example input and desired output pair for the given task, and the resulting state of the input sequence and the output buffer after executing the currently written program.

At each state, the agent has $N$ actions $= \{0, 1, ..., N-1\}$, each corresponding to writing that number into the next location in memory. The memory is initially filled with zeroes. After an action is chosen, the number is added to memory, and the currently written program is executed to determine the states of the input and output on the example test as well as other (hidden) test cases. If all test cases succeed, the episode terminates after one more (irrelevant) action. On the other hand, if the agent reaches the end of writable memory ($N - 3$ actions), the episode terminates unsuccessfully.

The Negate Positives task expects a program which receives positive integers and outputs that input, expect negated. This task are very simple, since it can be done with a single instruction: Reading from \texttt{@IN} will subtract the next input from whichever address to which we write. If we write to \texttt{@OUT}, which reads as $0$, we are effectively writing $0 - v_i = -v_i$. For Negate Positives, this is exactly what is required, so the only remaining challenge is how to loop back to the start. Luckily, since we know the input is always positive, the output will be negative (or zero), so we always jump. Thus, if we choose the jump address to be $0$, we can also loop to the start in the same instruction.

$(\texttt{@OUT}, \texttt{@IN}, 0, ...)$ in memory forms a solution to the task, and since the memory is initialized with zeroes, it only requires the agent to write $2$ words, meaning only $2$ actions.

In code, a solution to Negate Positives looks like this:
\begin{verbatim}
@start:
subleq @OUT @IN @start
\end{verbatim}
Which the agent would write as:
\begin{verbatim}
(N-2, N-3, )
\end{verbatim}

In the Identity task, we require a program which outputs the input unchanged. This is slightly more difficult that Negate Positives, because an instruction in {\sc subleq} always subtracts, so we need at least two instruction to undo the negation. This can be done by storing the return of \texttt{@IN} in some auxiliary word, and then writing that value to output. Let us denote the address at which we store the value temporarily as \texttt{@x}. The program that solves Identity could begin $(\texttt{@x}, \texttt{@IN}, 3, \texttt{@OUT}, \texttt{@x}, ...)$. Note the $3$ at address $2$ is needed to progress to the second instruction regardless of whether the input is positive or negative. This program would work for the first input, but we must produce a program which loops. Thus, we still need to erase the value stored at address \texttt{@x} (to prepare it for the next write), and we must loop to the start. These two things can be done in the same instruction. We arrive at a program like $(\texttt{@x}, \texttt{@IN}, 3, \texttt{@OUT}, \texttt{@x}, 6, \texttt{@x}, \texttt{@x}, 0, ...)$. The last instruction $(\texttt{@x}, \texttt{@x}, 0)$ clears \texttt{@x} and jumps to address zero (since $w_{\texttt{@x}} \leftarrow w_{\texttt{@x}} - w_{\texttt{@x}} = 0$ which $\le 0$). Again, we required the constant $6$ at address $5$ to unconditionally continue to the next address. In general, an agent would need $8$ actions to solve Identity, where it needs to pick \texttt{@IN}, \texttt{@OUT}, $3$, and $6$ specifically, and it needs to make sure that all of \texttt{@x} are the same. We say in general, because there is actually a clever solution when we choose $@x = 0$. In that case, the program becomes $(0, \texttt{@IN}, 3, \texttt{@OUT}, 0, 6, 0, 0, 0, ...)$ which only requires the agent to write $6$ words, since the memory is initially filled with zeroes.

In code, a general solution to Identity looks like this (left in human syntax, right as the agent writes, the 9s can be any constant $\ge 9$, $< N$):

\begin{verbatim}
@start:
subleq @x @IN          ; (9, N-3, 3)
subleq @OUT @x         ; (N-2, 9, 6)
subleq @x @x @start    ; (9, 9, )
@x: .data 0            ; (, ...)
\end{verbatim}

While the shortest (known) solution (in terms of non-zero bytes written) looks like this:

\begin{verbatim}
@start:
subleq @x:@x @IN       ; (0, N-3, 3)
subleq @OUT @x         ; (N-2, 0, 6)
subleq @x @x @start    ; (, , )
\end{verbatim}

\section{Implementation Details for Deep Sea}
\label{appendix_implementation}
Our implementation for the Deep Sea agents is based in the framework of \cite{efficientzero}. 
Below, we detail the implementation details unique to E-A/MZ and A/MZ+UBE in  Deep Sea.

% Targets
\subsection{Targets}
\label{appendix_implementation_targets}
Value, policy and reward targets were all computed as in MZ \cite{MuZero}.
UBE targets were computed in an n-step manner:
\begin{align}
        u_{target}(s_t) = \sum_{i = 0}^{n-1}\gamma^{2i} 
     \eta(s_{t+i}, a_{t+i}) + 
     \gamma^{2n}
     \V \,[\hat V_{\hat M}^\pi(s_{t+n})]
\end{align}
Where $\eta$ is the RND / exact or hash count-based novelty estimators.
To guarantee that the UBE estimates remains sufficiently optimistic, the value-uncertainty bootstrap $ \V \,[\hat V_{\hat M}^\pi(s_{t+n})] $ was computed in one of two ways:
\begin{enumerate}
    \item \textit{Root targets}:
    The uncertainty at the root of the EMCTS tree at state $s_{t+n}$. 
    \item \textit{Non-root targets}:
    \begin{align}
        \V \,[\hat V_{\hat M}^\pi(s_{t+n})] = \max_a \eta(s_{t+n}, a_{t+n}) + 
     \gamma^{2}u(s_{t+n+1}),
    \end{align}
\end{enumerate}
Where 
\begin{align}
    u(s_{t+n+1}) = \max \big (\hat u (s_{t+n+1}), \frac{1}{1-\gamma^2}\V[\hat R(s_{t+n},a_{t+n})] \big )
\end{align}

% Losses
\subsection{Losses}
\label{appendix_implementation_losses}
The original MZ algorithm uses three loss functions $ L_r, L_v, L_\pi $ for the reward, value and policy, respectively.
The gradients from the losses $ \Set L_r, \Set L_v, \Set L_\pi $ propagate through the learned transition model $f$ and are the only learning signal that is used to train the model.

For the anchored model (see Section~\ref{results}) we use an additional reconstruction loss:
\begin{align*}
	\Set L_{re} :=
            \smallfrac{1}{|\mathcal B|}\smallsum{t \in \mathcal B}{} \smallsum{k=0}{l-1}
                ||\hat s_t^k -  s_{t+k}||^2
\end{align*}
where $ \mathcal B \equiv \{s_t, a_t, r_t, s_{t+1}, a_{t+1}, \ldots, s_{t+l} \}_{t \in \mathcal B} $ is a training batch containing $ b $ trajectories of length $ l $ sampled from different episodes, and $ \hat s_t^k $ is the state predicted by the learned model.
To simplify model learning with the anchored model, the representation function $g$ that was used for the anchored model transforms the observations from 2 dimensional $(N,N)$ one-hot representations to 1 dimensional $(2N)$ representations where the first $N$ entries are a 1-hot vector representing the row and following $N$ entries are a 1-hot vector representing the column.
From this perspective, we can view the $\Set L_{re}$ loss that was used to train the anchored model as a consistency loss between the representation and the state prediction rather than a reconstruction loss. 

To estimate value-uncertainty at the leaves, we train a UBE function $u$ with a UBE loss $ \Set L_{u} $:
\begin{align*}
	\Set L_{u} :=
	       \smallfrac{1}{|\mathcal B|}\smallsum{t \in \mathcal B}{} \smallsum{k=0}{l-1}
                \phi(u^{\text{target}}_{t+k})^T \log \hat u_t^k
\end{align*}

The final loss is computed as:
\begin{align*}
    \Set L := \lambda_r \Set L_r + \lambda_v \Set L_v + \lambda_\pi \Set L_\pi + \lambda_u \Set L_u
\end{align*}
Where the coefficients $ \lambda_r, \lambda_v, \lambda_\pi, \lambda_u $ are used to weigh the relative effects the individual components of the loss have on the learned transition model $f$.
When $ \Set L_{re} $ was used (the anchored model in Section~\ref{results}), the model parameters of $f$ were affected only by $L_{re}$, through a second backwards pass.

% Different dynamics models
\subsection{Different Dynamics Models}
\label{appendix_dynamics_models}
We describe the three transition models used in \ref{results} in more detail.
The AlphaZero dynamics model is a true model of the dynamics of the environment, in the true state space of the environment.
When planning with this model local uncertainty is estimated with RND and value-uncertainty is estimated with UBE.
The MuZero model is a value-equivalent model in latent space. $g,f$ are learned by the agent during training from the value, policy, reward and UBE losses.
When planning with this model local uncertainty is estimated with state-visitation-counts (see \ref{appendix_planning_with_counts} and value-uncertainty is estimated with UBE.
The anchored-MuZero transition model trained only to predict the true transition dynamics of the environment through a reconstruction loss $L_{re}^k$ (see Appendix \ref{appendix_implementation_losses}). 
When planning with this model local uncertainty is estimated with RND and value-uncertainty is estimated with UBE.

% RND
\subsection{Planning with Random Network Distillation Based Epistemic Uncertainty}
\label{appendix_planning_with_RND}
In order to estimate the epistemic uncertainty of a transition $\eta(s,a)$, RND \citep{rnd} take as input state $s$ and action $a$ and computes $ L_2(\phi(s,a), \phi'(s,a)) $ between two neural networks $ \phi, \phi' $. 
$\phi'$ is kept stationary, while $\phi$ is trained with the same loss that evaluates the uncertainty.
When the planning is done with a true model, the agent has access to the true states $s_{t+k}$.
When the planning is done with the anchored model, the latent states outputted by the transition model $\hat{s}^k_t$ approximate the true states $s_{t+k}$ which allows us to use RND over $(\hat{s}^k_t, a_{t+k})$.
In both cases, RND is trained only over the observed transitions $(s_{t+k}, a_{t+k})$, not latent state representations $(\hat{s}^k_t, a_{t+k})$, to achieve the objective of yielding large RND prediction errors the further the latent state prediction $\hat{s}^k_t$ is from observed state $s_{t+k}$.

\subsection{Planning with Visitation-Counts Based Epistemic Uncertainty}
\label{appendix_planning_with_counts}
When planning with the abstracted model, we provide the agent with access to two additional mechanisms that are used only for local uncertainty estimation: the true model of the environment and a state-action visitation counter $C(s_t, a_t)$.
During planning, the true transition model follows the planning decisions $a_{t:t+k}$ and keeps track of the true state $s_{t+k}$.
When the agent evaluates the local uncertainty with transition $(\hat{s}^k_t, a_{t+k})$ the true model provides the matching true state $s_{t+k}$ to the visitation counter, which produces the local uncertainty based on the following formula:
\begin{align*}
    \eta(s_{t+k}, a_{t+k}) = \frac{1}{C(s_{t+k}, a_{t+k}) + \epsilon}
\end{align*}
Where $\epsilon > 0$ is a constant and $C(s_{t+k}, a_{t+k})$ counts the number of times the state action pair $ (s_{t+k}, a_{t+k}) $ has been observed in the environment.
This allows us to evaluate the abstracted-model agent in the presence of a reliable source of local uncertainty.
The leaf-value uncertainty $u(\hat{s}^k_t)$ (which is the dominating factor in visited areas of the state space, as $\eta(s_{t+k}, a_{t+k}) \to 0$ quickly in observed transitions) relies entirely on the learned UBE function $u$ which operates directly on latent states $\hat{s}^k_t$.

\subsection{Using UBE to Estimate Value-Uncertainty at the Leaves}
\label{appendix:ube_predictions}
It is essential for exploration that the epistemic uncertainty prediction is reliably high in unobserved areas of the state action space.
For this reason, a learned function $ \hat u \approx u $ may not be sufficient to detect that a state $ s_t $ has not been previously observed.
Instead, we use the following: 
\begin{align}
    \max \big (\hat u (s_t), \frac{1}{1-\gamma^2}\eta(s_t, \pi(s_t)) \big )
    \label{eq:ube_max}
\end{align}
If the uncertainty $\eta(s_t, \pi(s_t))$ for the transition $(s_t, \pi(s_t))$ is high, the uncertainty will be estimated as high regardless of the UBE prediction $u (s_t)$, and otherwise, either $u (s_t)$ is high or both are negligible.

\subsection{Separating Exploration from Exploitation}
Acting in the environment with a dedicated exploration policy can be expected to result in samples that are very off-exploitation-policy.
Learning from very off-policy data is known to cause instability in training even in off-policy agents.
To mitigate that, the EMCTS and only-UBE agents (see section \ref{results}) alternate between two types of training episodes: \textit{exploratory episodes} that follow an exploration policy throughout the episode (such as a policy generated by EMCTS with an exploratory planning objective), and \textit{exploitatory episodes} that follow the standard MuZero exploitation policy throughout the episode.
This enables us to provide the agent with quality exploitation targets to evaluate and train the value and policy functions reliably, while also providing a large amount of exploratory samples that explore the environment much more effectively and are more likely to efficiently search for high-reward interactions. 

In practice, rather than alternate between exploration and exploitation episodes we run a certain number of episodes in parallel, a certain portion of which are exploitatory and the rest are exploratory.
In our experiments the ratio was $50 / 50$.
To avoid learning a a separate prior-policy that may not be necessary in environments with small actions space, we set the policy prediction $ \pi(s_k) $ (see Equation~\ref{eq:PUCT}) to uniform over all actions, for all $ s_k $ during exploration episodes.
Dirichlet noise was not used to drive exploration in MCTS with the UBE and EMCTS agents, as any little amount of stochasticity in the policy can prevent the agent from reliably completing the one optimal trajectory.

\subsection{The A/MZ+UBE Agent}
\label{app:only_ube}
The A/MZ+UBE ablation agent uses MCTS to evaluate the $q$ value of actions in the same manner as A/MZ, and explores by taking the action $a_t$ that maximizes the combination of the Q-values approximated by MCTS, local uncertainty $\V\,[\hat{R}]$ and UBE head $u$: 
\begin{align}
\label{eq:only_ube}
    a_t \;=\; \argmax_a q_{\hat M}(s_0, a_t) + \beta \sqrt{\V\,[\hat{R}(s_0, a_t)] + \gamma^2 u(f(s_0,a_t))} .
\end{align}
$q_{\hat M}(s_0, a_t)$ are the values at the root of the regular MCTS tree after search.
The main difference between A/MZ+UBE and E-A/MZ is that in E-A/MZ the uncertainty $ \sqrt{\V\,[\hat{R}(s_0, a_t)] + \gamma^2 u(f(s_0,a_t))} $ estimated takes into account estimates from multiple different future trajectories, in the manner MCTS estimates the values $ q_{\hat M} $.

\section{Network Architecture \& Hyperparameters}
\label{appendix_network_arch}
\subsection{Hyperparameter Search}
Due to the large number of hyperparameters in the MuZero framework, our optimization process consisted of manual modifications to the hyperparameters used by \citet{efficientzero} for Deep Sea and \citet{koyamada2023pgx} for {\sc subleq} with the objective of achieving learning stability on the target environment with the simplest network architectures possible.
Two exceptions to this statement are the RND network architecture and scale, and the exploration parameter $\beta$.

The RND architecture was designed with the objective of reliably achieving small RND predictions over observed state-action pairs and large predictions over unobserved state-action pairs.
The RND scale was tuned with the objective of achieving local uncertainty measures for unobserved state-action pairs that are significantly larger than the minimum reward of Deep Sea.

The $\beta$ parameter in Deep Sea was tuned with the objective that the EMCTS and only-UBE agents will prioritize exploration of the environment over exploitation until the entire environment has been searched, and was tuned separately for every model.

For {\sc subleq} we chose $\beta = 1$.
We did not experiment with additional values of $\beta$. However, first, the values, UBE prediction, and state-novelty predictions are all bounded $\leq 1$, such that $\beta$ need not account in this case to the possibly arbitrary scales of UBE / the novelty estimator.
Second, to make the most out of Jax's naturally parallelized setup, each parallel episode explores with a different $\beta_i \leq \beta$, evenly spaced from $0$ to $\beta$.

\subsection{Network Architecture}
The functions $f, g, r, v, u, \pi, \psi, \psi'$ used fully connected DNNs of varying sizes.
The sizes of the hidden layers and output layers are specified in Table~\ref{table:network_arch} for Deep Sea.
For {\sc subleq} the FC network architecture constituted value, UBE, exploration policy $\pi_e$ and exploitation policy $\pi$ networks.
All networks used two hidden layers of size 256 with ReLu activations between hidden layers.
The value head used tanh activation on the last layer, the UBE head used a $0.5 (tanh(x) + 1)$ activation to bound the ube prediction between 0 and 1. The policy heads did not use any activation layers.

% Net archs
\begin{table}[t]
\centering
\caption{Network architecture hyperparameters, Deep Sea}
\begin{tabular}{c|cc|}
\cline{2-3}
                               & \multicolumn{2}{c|}{True Model}                                 \\ \hline
\multicolumn{1}{|c|}{Function} & \multicolumn{1}{c|}{Hidden Layers Sizes}    & Output Layer Size \\ \hline
\multicolumn{1}{|c|}{f}        & \multicolumn{1}{c|}{-}                      & -                 \\ \hline
\multicolumn{1}{|c|}{g}        & \multicolumn{1}{c|}{-}                      & -                 \\ \hline
\multicolumn{1}{|c|}{r}        & \multicolumn{1}{c|}{{[}256, 256{]}}         & 21                \\ \hline
\multicolumn{1}{|c|}{v}        & \multicolumn{1}{c|}{{[}256, 256{]}}         & 21                \\ \hline
\multicolumn{1}{|c|}{u}        & \multicolumn{1}{c|}{{[}256, 256{]}}         & 21                \\ \hline
\multicolumn{1}{|c|}{$\pi$}    & \multicolumn{1}{c|}{{[}256, 256{]}}         & 2                 \\ \hline
                               & \multicolumn{2}{c|}{Anchored Model}                             \\ \hline
\multicolumn{1}{|c|}{Function} & \multicolumn{1}{c|}{Hidden Layers Sizes}    & Output Layer Size \\ \hline
\multicolumn{1}{|c|}{f}        & \multicolumn{1}{c|}{{[}1024, 1024, 1024{]}} & 80                \\ \hline
\multicolumn{1}{|c|}{g}        & \multicolumn{1}{c|}{-}                      & -                 \\ \hline
\multicolumn{1}{|c|}{r}        & \multicolumn{1}{c|}{{[}256, 256{]}}         & 21                \\ \hline
\multicolumn{1}{|c|}{v}        & \multicolumn{1}{c|}{{[}256, 256{]}}         & 21                \\ \hline
\multicolumn{1}{|c|}{u}        & \multicolumn{1}{c|}{{[}256, 256{]}}         & 21                \\ \hline
\multicolumn{1}{|c|}{$\pi$}    & \multicolumn{1}{c|}{{[}256, 256{]}}         & 2                 \\ \hline
                               & \multicolumn{2}{c|}{Abstracted Model}                           \\ \hline
\multicolumn{1}{|c|}{Function} & \multicolumn{1}{c|}{Hidden Layers Sizes}    & Output Layer Size \\ \hline
\multicolumn{1}{|c|}{f}        & \multicolumn{1}{c|}{{[}1024, 1024, 1024{]}} & 100               \\ \hline
\multicolumn{1}{|c|}{g}        & \multicolumn{1}{c|}{{[}512, 512{]}}         & 100               \\ \hline
\multicolumn{1}{|c|}{r}        & \multicolumn{1}{c|}{{[}128, 128{]}}         & 21                \\ \hline
\multicolumn{1}{|c|}{v}        & \multicolumn{1}{c|}{{[}128, 128{]}}         & 21                \\ \hline
\multicolumn{1}{|c|}{u}        & \multicolumn{1}{c|}{{[}128, 128, 128{]}}    & 21                \\ \hline
\multicolumn{1}{|c|}{$\pi$}    & \multicolumn{1}{c|}{{[}128, 128{]}}         & 2                 \\ \hline
                               & \multicolumn{2}{c|}{RND network architecture}                   \\ \hline
\multicolumn{1}{|c|}{Function} & \multicolumn{1}{c|}{Hidden Layers Sizes}    & Output Layer Size \\ \hline
\multicolumn{1}{|c|}{$\psi$}   & \multicolumn{1}{c|}{{[}1024, 1024{]}}       & 512               \\ \hline
\multicolumn{1}{|c|}{$\psi'$}  & \multicolumn{1}{c|}{[512]}                    & 512               \\ \hline
\end{tabular}
\label{table:network_arch}
\end{table}

\subsection{Deep Sea Hyperparameter Configuration}
We detail the full set of hyperparameters in Tables~\ref{table:shared_hps} and~\ref{table:unique_hps} for Deep Sea.
For the BDQN baseline, we used the default implementation in \url{https://github.com/deepmind/bsuite}, with ensemble size of 10 and matching batch size to EMCTS: number of unroll steps times batch size $ 5 \cdot 256 = 1230 $.
Otherwise, the default hyper parameters were used.

\begin{table}[t]
\small
\setlength\tabcolsep{0pt}
\centering
\caption{Hyperparameters used in the Deep Sea experiments}
\begin{tabular}{|c|c|c|}
\hline
Parameter                          & Setting                            & Comment \\ \hline
Stacked Observations               & 1                                  &         \\ \hline
$\gamma$                           & 0.995                              &         \\ \hline
Number of simulations in MCTS      & 50                                 &         \\ \hline
Dirichlet noise ratio ($\xi$)     & 0.3                                &         \\ \hline
Root exploration fraction          & 0                                  &         \\ \hline
Batch size                         & 256                                &         \\ \hline
Learning rate                      & 0.0005                             &         \\ \hline
Optimizer                          & Adam \citep{ADAM} &         \\ \hline
Unroll steps $l$                   & 5                                  &         \\ \hline
Value target TD steps ($n_v$)      & 5                                  &         \\ \hline
UBE target TD steps ($n_u$)        & 1                                  &         \\ \hline
value support size                 & 21                                 &         \\ \hline
UBE support size                   & 21                                 &         \\ \hline
Reward support size                & 21                                 &         \\ \hline
Reanalyzed policy ratio       & 0.99          & See \citep{efficientzero}                                                \\ \hline
Prioritized sampling from the replay     & True          & \begin{tabular}[c]{@{}c@{}} See \citep{MuZero} \\ Appendix G \end{tabular} \\ \hline
Priority exponent ($\alpha$)  & 0.6           & \begin{tabular}[c]{@{}c@{}} See \citep{MuZero} \\ Appendix G \end{tabular} \\ \hline
Priority correction ($\beta_p$) & $ 0.4 \to 1 $ & \begin{tabular}[c]{@{}c@{}} See \citep{MuZero} \\ Appendix G \end{tabular} \\ \hline
Evaluation episodes                & 8                                 &         \\ \hline
Min replay size for sampling       & 300                                &         \\ \hline
Self-play network updating inerval & 5                                  &         \\ \hline
Target network updating interval   & 10                                 &         \\ \hline
\end{tabular}
\label{table:shared_hps}
\end{table}

\begin{table}[t]
\small
\setlength\tabcolsep{1.4mm}
\centering
\caption{Specific  for models and agents}
\begin{tabular}{|c|ccccccccc|}
\hline
\multirow{3}{*}{Parameter} &
  \multicolumn{9}{c|}{Setting} \\ \cline{2-10} 
 &
  \multicolumn{3}{c|}{True Model} &
  \multicolumn{3}{c|}{Abstracted Model} &
  \multicolumn{3}{c|}{Anchored Model} \\ \cline{2-10} 
 &
  \multicolumn{1}{c|}{EMCTS} &
  \multicolumn{1}{c|}{UBE} &
  \multicolumn{1}{c|}{Uninf.} &
  \multicolumn{1}{c|}{EMCTS} &
  \multicolumn{1}{c|}{UBE} &
  \multicolumn{1}{c|}{Uninf.} &
  \multicolumn{1}{c|}{EMCTS} &
  \multicolumn{1}{c|}{UBE} &
  Uninf. \\ \hline
\begin{tabular}[c]{@{}c@{}}Training steps /\\ environment \\ interactions\end{tabular} &
  \multicolumn{1}{c|}{45K} &
  \multicolumn{1}{c|}{45K} &
  \multicolumn{1}{c|}{45K} &
  \multicolumn{1}{c|}{35K} &
  \multicolumn{1}{c|}{35K} &
  \multicolumn{1}{c|}{35K} &
  \multicolumn{1}{c|}{45K} &
  \multicolumn{1}{c|}{45K} &
  45K \\ \hline
\begin{tabular}[c]{@{}c@{}}Reward loss \\ weight $\lambda_r$\end{tabular} &
  \multicolumn{1}{c|}{1} &
  \multicolumn{1}{c|}{1} &
  \multicolumn{1}{c|}{1} &
  \multicolumn{1}{c|}{1} &
  \multicolumn{1}{c|}{1} &
  \multicolumn{1}{c|}{1} &
  \multicolumn{1}{c|}{1} &
  \multicolumn{1}{c|}{1} &
  1 \\ \hline
\begin{tabular}[c]{@{}c@{}}Value-loss \\ weight $\lambda_v$\end{tabular} &
  \multicolumn{1}{c|}{0.5} &
  \multicolumn{1}{c|}{0.5} &
  \multicolumn{1}{c|}{0.5} &
  \multicolumn{1}{c|}{0.5} &
  \multicolumn{1}{c|}{0.5} &
  \multicolumn{1}{c|}{0.5} &
  \multicolumn{1}{c|}{0.5} &
  \multicolumn{1}{c|}{0.5} &
  0.5 \\ \hline
\begin{tabular}[c]{@{}c@{}}Policy-loss \\ weight $\lambda_\pi$\end{tabular} &
  \multicolumn{1}{c|}{0.5} &
  \multicolumn{1}{c|}{0.5} &
  \multicolumn{1}{c|}{0.5} &
  \multicolumn{1}{c|}{0.5} &
  \multicolumn{1}{c|}{0.5} &
  \multicolumn{1}{c|}{0.5} &
  \multicolumn{1}{c|}{0.5} &
  \multicolumn{1}{c|}{0.5} &
  0.5 \\ \hline
\begin{tabular}[c]{@{}c@{}}UBE-loss \\ weight $\lambda_u$\end{tabular} &
  \multicolumn{1}{c|}{0.125} &
  \multicolumn{1}{c|}{0.125} &
  \multicolumn{1}{c|}{-} &
  \multicolumn{1}{c|}{0.25} &
  \multicolumn{1}{c|}{0.25} &
  \multicolumn{1}{c|}{-} &
  \multicolumn{1}{c|}{0.125} &
  \multicolumn{1}{c|}{0.125} &
  - \\ \hline
RND scale &
  \multicolumn{1}{c|}{1.0} &
  \multicolumn{1}{c|}{1.0} &
  \multicolumn{1}{c|}{-} &
  \multicolumn{1}{c|}{-} &
  \multicolumn{1}{c|}{-} &
  \multicolumn{1}{c|}{-} &
  \multicolumn{1}{c|}{0.001} &
  \multicolumn{1}{c|}{0.001} &
  - \\ \hline
\begin{tabular}[c]{@{}c@{}}Root based \\ targets\end{tabular} &
  \multicolumn{1}{c|}{False} &
  \multicolumn{1}{c|}{False} &
  \multicolumn{1}{c|}{False} &
  \multicolumn{1}{c|}{True} &
  \multicolumn{1}{c|}{True} &
  \multicolumn{1}{c|}{True} &
  \multicolumn{1}{c|}{False} &
  \multicolumn{1}{c|}{False} &
  False \\ \hline
\begin{tabular}[c]{@{}c@{}}Disabled \\ policy \\ in exploration\end{tabular} &
  \multicolumn{1}{c|}{True} &
  \multicolumn{1}{c|}{True} &
  \multicolumn{1}{c|}{False} &
  \multicolumn{1}{c|}{True} &
  \multicolumn{1}{c|}{True} &
  \multicolumn{1}{c|}{False} &
  \multicolumn{1}{c|}{True} &
  \multicolumn{1}{c|}{True} &
  False \\ \hline
\begin{tabular}[c]{@{}c@{}}Number of \\ parallel \\ episodes\end{tabular} &
  \multicolumn{1}{c|}{2} &
  \multicolumn{1}{c|}{2} &
  \multicolumn{1}{c|}{2} &
  \multicolumn{1}{c|}{2} &
  \multicolumn{1}{c|}{2} &
  \multicolumn{1}{c|}{2} &
  \multicolumn{1}{c|}{2} &
  \multicolumn{1}{c|}{2} &
  2 \\ \hline
\begin{tabular}[c]{@{}c@{}}Out of are\\ exploration \\ episodes\end{tabular} &
  \multicolumn{1}{c|}{1} &
  \multicolumn{1}{c|}{1} &
  \multicolumn{1}{c|}{-} &
  \multicolumn{1}{c|}{1} &
  \multicolumn{1}{c|}{1} &
  \multicolumn{1}{c|}{-} &
  \multicolumn{1}{c|}{1} &
  \multicolumn{1}{c|}{1} &
  - \\ \hline
\begin{tabular}[c]{@{}c@{}}Exploration \\ coefficient $\beta$\end{tabular} &
  \multicolumn{1}{c|}{10} &
  \multicolumn{1}{c|}{10} &
  \multicolumn{1}{c|}{-} &
  \multicolumn{1}{c|}{1} &
  \multicolumn{1}{c|}{1} &
  \multicolumn{1}{c|}{-} &
  \multicolumn{1}{c|}{10} &
  \multicolumn{1}{c|}{10} &
  - \\ \hline
\begin{tabular}[c]{@{}c@{}}Dirichlet noise \\ magnitude $\rho$ \end{tabular} &
  \multicolumn{1}{c|}{\begin{tabular}[c]{@{}c@{}} 0\end{tabular}} &
  \multicolumn{1}{c|}{\begin{tabular}[c]{@{}c@{}} 0\end{tabular}} &
  \multicolumn{1}{c|}{0.25} &
  \multicolumn{1}{c|}{\begin{tabular}[c]{@{}c@{}} 0\end{tabular}} &
  \multicolumn{1}{c|}{\begin{tabular}[c]{@{}c@{}} 0\end{tabular}} &
  \multicolumn{1}{c|}{0.25} &
  \multicolumn{1}{c|}{\begin{tabular}[c]{@{}c@{}} 0\end{tabular}} &
  \multicolumn{1}{c|}{\begin{tabular}[c]{@{}c@{}} 0\end{tabular}} &
  0.25 \\ \hline
\end{tabular}
\label{table:unique_hps}
\end{table}

\subsection{SUBLEQ Hyperparameter Configuration}
The {\sc subleq} E-/AZ agents are implemented in Jax, based in the implementation of \cite{koyamada2023pgx}.
MCTS parameters used the defaults provided by \cite{deepmind2020jax}.
Detailed hyperparameter configuration below:
\begin{enumerate}
    \item hash: XXHash, 32bit
    \item number of parallel episodes = 128   
    \item number of E/MCTS simulations = 32
    \item batch size = 4096
    \item number of times each frame appears in training in expectation = 4
    \item discount = 0.97
    \item replay buffer size = 200,000
    \item priority exponent of prioritized replay buffer = 0.6
    \item learning rate = 0.001
    \item populate replay buffer for N frames before starting training, N = 5000
    \item run evaluation episode every N frames, N = 20480
    \item number of parallel evaluation episodes = 32
\end{enumerate}

\end{document}